\documentclass{article}

\usepackage{arxiv}

\usepackage[english]{babel}

\usepackage{graphicx}
\usepackage{amssymb,amsmath,bm}
\usepackage{textcomp}
\usepackage{booktabs}
\usepackage[textfont=it,tableposition=top]{caption}
\usepackage{siunitx}
\usepackage{xspace}
\usepackage{url}
\usepackage{lipsum}
\usepackage{enumitem}
\usepackage{xkeyval}
\usepackage{calc}

\usepackage{float}  
\usepackage{xcolor} 
\usepackage{soul}  
\usepackage{authblk}  
\usepackage{csquotes}
\usepackage[style=ieee,backend=biber,maxnames=10,minnames=1]{biblatex}

\usepackage{hyperref}
\hypersetup{colorlinks=true, allcolors=blue}

\renewcommand{\headeright}{}
\renewcommand{\undertitle}{}
\renewcommand{\shorttitle}{}

\newcommand{\signa}{\textsuperscript{a}}
\newcommand{\signb}{\textsuperscript{b}}
\newcommand{\signc}{\textsuperscript{c}}
\newcommand{\nosign}{\phantom{\textsuperscript{c}}}

\title{Whisper-LM: Improving ASR Models with Language Models for Low-Resource Languages}

\makeatletter
\renewcommand\AB@affilsepx{: \protect\Affilfont}  
\makeatother

\newbox{\orcid}\sbox{\orcid}{\includegraphics[scale=0.06]{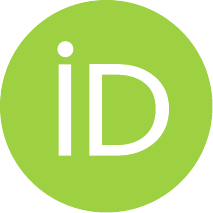}}
\author{%
	\href{https://orcid.org/0009-0001-9309-8507}{\usebox{\orcid}\hspace{1mm}Xabier de Zuazo\thanks{\texttt{xabier.dezuazo@ehu.eus}}}%
}
\author{%
	\href{https://orcid.org/0000-0003-3804-4984}{\usebox{\orcid}\hspace{1mm}Eva Navas\thanks{\texttt{eva.navas@ehu.eus}}}%
}
\author{%
	\href{https://orcid.org/0000-0001-7282-2765}{\usebox{\orcid}\hspace{1mm}Ibon Saratxaga\thanks{\texttt{ibon.saratxaga@ehu.eus}}}%
}
\author{%
	\href{https://orcid.org/0000-0003-4447-7575}{\usebox{\orcid}\hspace{1mm}Inma Hernáez Rioja\thanks{\texttt{inma.hernaez@ehu.eus}}}%
}
\affil{HiTZ - University of the Basque Country - UPV/EHU, Spain}

\date{March, 2025}

\begin{document}
\maketitle

\begin{abstract}
    Automatic speech recognition systems have undoubtedly advanced with the integration of multilingual and multitask models such as Whisper, which have shown a promising ability to understand and process speech across a wide range of languages. Despite their robustness, these models often fall short in handling the linguistic distinctions of minority languages. This study addresses this gap by integrating traditional and novel language models with fine-tuned Whisper models to raise their performance in less commonly studied languages. Through rigorous fine-tuning and evaluation across multiple datasets, we demonstrate substantial improvements in word error rate, particularly in low-resource scenarios. Our approach not only does take advantage of the extensive data Whisper was pre-trained on, but also complements its linguistic adaptability by incorporating language models. We obtained improvements up to 51\% for in-distribution datasets and up to 34\% for out-of-distribution sentences using statistical language models, while large language models provided moderate but consistently robust improvement across diverse linguistic contexts. The findings reveal that, while the integration reliably benefits all model sizes, the extent of improvement varies, highlighting the importance of optimized language model parameters. Finally, we emphasize the importance of selecting appropriate evaluation parameters when reporting the results using transformer-based ASR models. In summary, this research clears the way for more inclusive ASR technologies that perform better across languages by enriching their linguistic knowledge. For further implementation details of this study, the technical documentation and source code are available at \url{http://www.github.com/hitz-zentroa/whisper-lm}.\footnote{This work has been submitted to the IEEE for possible publication. Copyright may be transferred without notice, after which this version may no longer be accessible.}
\end{abstract}

\section{Introduction}

In order to enable computers to comprehend and analyze human speech, automatic speech recognition (ASR) technology attempts to convert spoken language into text. The recent development and use of unsupervised and weakly supervised learning techniques have significantly expedited progress in ASR. One of these systems, Whisper, is a multilingual and multitask ASR model that has shown impressive performance in processing and recognizing speech in a wide range of languages and domains~\cite{radford2022}. This innovation is further strengthened by the growing volume of pre-training data, which includes up to 680,000 hours of tagged audio to enhance generalization and robustness across datasets.

However, despite its effectiveness, Whisper and similar models often struggle with linguistic characteristics and grammatical peculiarities of low-resource languages, necessitating further refinement to optimize performance. This motivates the integration of language models (LMs), which have historically played a vital role in enhancing linguistic coherence and context handling in ASR systems. By embedding language models, we aim to bridge the gap between the phonetic richness captured by Whisper's encoder and the unique syntactic and semantic structures these languages exhibit.

Beyond closing the linguistic gap, a key achievement of this work is to improve the out-of-distribution (OOD) robustness, as ASR applications often exhibit worse performance when confronted with unseen or real-world data. To that end, we integrate language models into Whisper not through prompting but by merging their internal scores at inference time. This method augments Whisper’s outputs with an external language prior and effectively mitigates the domain-shift issues. In addition, we propose the Effective Robustness of Relative Error Reduction (noted ERER), a novel metric that quantifies how consistently the performance of a model scales from in-distribution to OOD scenarios. As our experiments show, both n-gram and large language models can substantially improve OOD performance under this framework, offering a more robust path toward more powerful ASR models across a variety of domains.

Additionally, the complexity and variability of real-world applications highlight the need to critically evaluate the impact of various model configuration parameters, such as those used at evaluation time. Despite not being very studied, these parameters considerably influence the ASR models' reported results when extended to novel or challenging linguistic environments. This paper seeks to methodically explore these aspects, exhaustively analyzing how different evaluation-time settings within the Whisper framework affect model performance across diverse datasets.

Our primary contribution is to demonstrate improved accuracy and robustness in speech recognition for low-resource languages through the integration of language models with weakly supervised acoustic ASR systems. This includes discarding possible data leakage by a sentence-level overlap analysis. We also provide a detailed exploration of the evaluation parameters that prominently affect the final results. This study marks the first documented instance of integrating Whisper with different language models, including an exhaustive comparison of their respective contributions. This methodology incorporates rigorous fine-tuning experiments supplemented by language model integration, offering insights into the synergistic combination of contemporary deep learning ASR models with both traditional and novel language models.

\section{Related Work}

The integration of language models into ASR systems, while not new, remains highly relevant in the era of deep learning-driven speech recognition technologies. The profound advancements in ASR, notably through deep recurrent neural networks~\cite{graves2013}, attention mechanisms~\cite{chan2015}, and transformer models~\cite{zhang2020}, have pushed the limits of what these systems can achieve. These advances underscore the evolution of ASR from traditional methods to modern approaches that integrate complex neural network architectures, addressing challenges in various applications ranging from industrial sectors to military and cultural applications~\cite{karmakar2021, hinton2012, zhang2017, alharbi2021, li2022}, and even more sophisticated fields like cognitive neuroscience~\cite{defossez2023, millet2024, antonello2024, sato2024}. In fact, multiple studies demonstrate that ASR systems can surpass human-level accuracy in certain environments, which prepares the path for their extensive deployment in real-world applications~\cite{amodei2016, baevski2020}.

Research on robust ASR systems often involves semi-supervised and self-supervised learning approaches, which use unlabeled data to improve learning efficacy~\cite{baevski2020, schneider2019, baevski2019, hsu2021, chen2022, conneau2021, babu2022, yang2021, mohamed2022, chang2021, peng2022, chung2021, baevski2022, liu2022}. These methods have proven particularly effective in escalating the ASR systems to handle multiple languages and dialects, which is essential to create inclusive technologies that cover the global linguistic diversity~\cite{narayanan2018, chan2021, zhang2022_2, li2021, zhang2023, communication2023_1, communication2023_2}. There are even some initial efforts to introduce semantic information into self-supervised encoders using unlabeled data~\cite{xu2022}.

Furthermore, the field has explored various strategies to adapt ASR systems to low-resource settings, often leveraging transfer learning and data augmentation techniques~\cite{wang2015, khare2021, meng2021}. A remarkable example of integrating LLMs with ASR models for low-resource languages is the work by Vasquez-Correa et al.~\cite{vasquezcorrea2024}, who explored the use of language models to select between different transcriptions provided by Zipformer ASR systems~\cite{yao2024}. This represents an initial, albeit shallow, integration of LLMs in ASR, suggesting promising future pathways for deeper synergies between these technologies. These strategies are vitally important for improving speech recognition accuracy in languages that lack extensive training corpora.

On the other hand, more recent techniques use weakly supervised pre-training methods to exploit large datasets with noisy, imprecise labeling~\cite{zhou2018}. This approach is particularly advantageous when labeled data is scarce or of low quality, which is often true with multilingual speech transcriptions. Whisper models, developed using weakly supervised pre-training, are excellent in handling multilingual speech recognition tasks. This competence positions Whisper models as prime candidates for further improvement through the integration of language models, especially to boost performance in low-resource languages.

A considerable amount of research has also been committed to the application of Whisper for various speech recognition tasks, highlighting its adaptability and the extensive scope of its training data~\cite{radford2022, gandhi2023, shao2024}. Along the way, multiple approaches and techniques to fine-tune Whisper-based models have been developed~\cite{zhou2017, sicard2023, timmel2024, liu2024, liu2024_2, xie2024}, and a more exhaustive evaluation of the models has been performed~\cite{gandhi2022}. In addition, the challenge of enhancing ASR performance for low-resource languages has been addressed by integrating Whisper with techniques to mitigate the impact of data imbalance, such as weighted cross-entropy for language discrimination~\cite{pineiro2024_1}, or by encoding the structured information into a sequence to allow multitask learning~\cite{pineiro2024_2}.

In this rich context of ongoing research and development, our work contributes to a deeper understanding of language model integration within the Whisper framework and its impact on ASR performance across diverse linguistic environments. The work is heavily based on previous approaches to integrate language models with neuronal ASR models~\cite{baevski2020,  gulati2020}. Our method leverages Whisper's foundational strengths, augmenting them with linguistic models to augment its performance and robustness in multiple languages. Specifically, we focused our contribution on four Iberian languages: three low-resource languages (Basque, Galician, and Catalan) and one high-resource language (Spanish)~\cite{gaspari2023}. In this regard, for our specific case, the Catalan language should be considered more a medium-sized language being more present online~\cite{gerrand2019, villegas2023} and, indeed, it has considerably higher representation in both the pre-training and fine-tuning datasets used here.

Besides, efforts to reproduce and improve Whisper models through open-source platforms and public datasets have gained traction, with initiatives like ESPnet~\cite{watanabe2018} and subsequent improvements in Whisper-style model training~\cite{peng2023, peng2024}. These efforts align with the open science movement, which promotes the transparency and accessibility of advanced models and training pipelines~\cite{zhang2022, liu2023, touvron2023_1, touvron2023_2, grattafiori2024, grattafiori2024}. In line with these open science principles, all of our material is released with an open-source license for free use and to encourage further research.

\section{Methodology}

This section outlines the methodologies used to improve the performance of multilingual Whisper models. We detail a structured fine-tuning process to enhance Whisper's performance across multiple languages. We then describe how language models are integrated at inference time to further increase accuracy and robustness. This includes an analysis of sentence leakage, and an ablation study to assess the impact of various evaluation parameters.

\subsection{Fine-Tuning Process}

We employed a standard fine-tuning process to adapt the multilingual Whisper models to improve performance in specific languages. Each model size, ranging from Tiny to Large-V3, was individually fine-tuned using the Hugging Face's community Whisper Fine-Tuning Event code\footnote{\label{fnot1} \url{https://github.com/huggingface/community-events}}. Fine-tuning was carried out on separate models for four distinct languages and seven different model sizes. This involved initializing the models with the pre-trained multilingual Whisper models provided by OpenAI (vanilla) and then fine-tuning them using the corresponding language subset from the Common Voice version 13.0 dataset~\cite{ardila2020}, including both the train and validation splits. This practice, recommended in the documentation for low-resource languages\footnote{\url{https://github.com/huggingface/community-events/tree/main/whisper-fine-tuning-event\#data}}, provides a larger training corpus than using only the train partition. However, it may lead to slightly optimistic results in the fine-tuning scores on the Common Voice dataset. In spite of that, as our study is focused on later language model integration benefits, this will not affect our final analysis. Ultimately, we ended with one fine-tuned model per language and size.

For homogeneity and fairness, parameters shared between languages were chosen based on the original scripts, and some adjustments derived from preliminary tests to improve performance while ensuring efficient training times. The learning rate is one of the most critical hyperparameters for Whisper model fine-tuning. Following the advice of Jong Wook Kim\footnote{\url{https://github.com/vasistalodagala/whisper-finetune/blob/master/README.md\#hyperparameter-tuning}}, one of the authors of the Whisper paper, we started with a learning rate that is approximately 40 times lower than the pre-training rate. Then, we adjusted the final learning rate based on initial training tests on the datasets and languages we used. The main idea was to use hyperparameters that worked well enough across languages to homogenize the fine-tuning processes and have an equivalent cross-language evaluation. These were the final learning rates used for each model size: Tiny models used a learning rate of \(3.75 \cdot 10^{-5}\), Base models \(2.5 \cdot 10^{-5}\), and Small to Large-V3 models \(1 \cdot 10^{-5}\). Based on our experience, starting with a learning rate of \(1 \cdot 10^{-5}\) gives better results with the largest models. Besides, we employed a linear learning rate scheduler with a warm-up phase of 500 steps, which is essential for stabilizing the training process early on. For all model sizes, the optimizer used was AdamW, with betas set to (0.9, 0.999) and epsilon to \(1 \cdot 10^{-08}\)~\cite{loshchilov2018}. The batch sizes were set based on our GPU's capacity, ranging from 256 for Tiny models to 32 for Large-V3 models, with gradient accumulation set to 2 for the largest models to compensate for the smaller batch size. Evaluation batch sizes were set correspondingly, from 128 for Tiny to 16 for Large-V3 models. Fine-tuning steps were adjusted to the complexity and size of the models: 5,000 steps for small models and up to 20,000 for the largest models. This ensured that each model was adequately trained to converge to optimal performance, reaching a good improvement plateau during training. A detailed summary of the hyperparameters used for fine-tuning each model size is provided in Table~\ref{tab:fine_tuning_hparams}.

\begin{table}[htp]
\centering
\caption{Summary of fine-tuning hyperparameters for the vanilla Whisper models.}
\label{tab:fine_tuning_hparams}
\begin{tabular}{l|rrrrr}
\hline
\textbf{Model Size} & \multicolumn{1}{c}{\textbf{Learning Rate}} & \multicolumn{1}{c}{\textbf{Batch Size}} & \multicolumn{1}{c}{\textbf{Eval Batch Size}} & \multicolumn{1}{c}{\textbf{Gradient Acc.}} & \multicolumn{1}{c}{\textbf{Steps}} \\ \hline
Tiny         & \(3.75 \cdot 10^{-5}\)  & 256 & 128 & 1 &  5,000 \\
Base         & \(2.5 \cdot 10^{-5}\)   & 128 & 64  & 1 &  5,000 \\
Small        & \(1 \cdot 10^{-5}\)     & 64  & 32  & 1 &  5,000 \\
Medium       & \(1 \cdot 10^{-5}\)     & 64  & 32  & 1 & 10,000 \\
Large-V1..V3 & \(1 \cdot 10^{-5}\)     & 32  & 16  & 2 & 20,000 \\
\hline
\end{tabular}
\end{table}

\subsection{Using Whisper with N-Gram Language Models}

Integrating n-gram language models with Whisper at inference time improves accuracy by combining the neural model's output with statistical language patterns captured by n-grams. This integration involves adjusting the log probabilities from Whisper's output based on the probabilities of the n-gram model.

The core of our integration involves modifying Whisper's beam search process. Precisely, we adjust the log probabilities of candidate tokens generated by Whisper using scores from a statistical language model. Equation~\ref{eq:whisper_lm} governs this adjustment, which has been previously used in the literature, including for Deep Speech models~\cite{hannun2014,amodei2016}.

\begin{equation} \label{eq:whisper_lm}
    Q(c|x) = \log(P_{\text{acoustic}}(c|x)) + \alpha \cdot \log(P_{\text{LM}}(c)) + \beta \cdot \text{word\_count}(c)
\end{equation}

In this equation, \(Q(c|x)\) represents the unnormalized log probability (or score) of the sequence \(c\) of candidate tokens, conditioned on the audio input \(x\). \(\log(P_{\text{acoustic}}(c|x))\) is the sequence's log probability from Whisper (acoustic model) given the audio input \(x\), \(\log(P_{\text{LM}}(c))\) is the log probability from the n-gram model, and \(\text{word\_count}(c)\) counts the words in \(c\). The parameters \(\alpha\) and \(\beta\), which scale the language model's and the sentence length's influence, respectively, are optimized for each language using a systematic approach detailed below. Notably, this equation is applied only at word boundaries. In addition, to prevent premature evaluation of short sequences, we introduce a length threshold based on the shortest recordings in the datasets. Specifically, language model scoring is applied only to sequences containing at least four Whisper tokens. These measures ensure that the language model's influence is applied when sufficient contextual information is present, enhancing the reliability of our results and avoiding repetitions and hallucinations.

As the n-gram language model, we selected KenLM for its ability to work successfully with an enormous linguistic corpus and for using Kneser-Ney smoothing to obtain relatively low perplexity. Furthermore, KenLM's efficient querying of n-gram probabilities is essential to maintain performance in ASR applications~\cite{heafield2011, heafield2013}. Based on previous work~\cite{hannun2014, amodei2016, tyers2021, babu2022}, we will restrict our approach to 5-gram language models for simplicity and to include a long enough linguistic context.

Concerning the parameter values to use, proper optimization of the \(\alpha\) and \(\beta\) in Equation~\ref{eq:whisper_lm} is indispensable to integrate the Whisper neural model with it smoothly. We employed a Bayesian optimization approach using the Optuna framework~\cite{akiba2019} for this optimization process. This process involves optimizing the parameters \(\alpha\) and \(\beta\) to minimize the Character Error Rate (CER)~\cite{klakow2002} or the Word Error Rate (WER)~\cite{wang2003}, depending on the characteristics of the language. For our optimization, Optuna uses its default single-objective sampler, the Tree-structured Parzen Estimator (TPESampler)~\cite{shuhei2023}. The TPESampler models the probability distribution of parameters using two Gaussian Mixture Models, one for parameters leading to better objective values and another for the rest. The sampler then selects new parameters by maximizing the ratio of these two models, concentrating the search on areas of the parameter space that lead to improved performance outcomes. For the search process to be more feasible, we have constrained the range of values of \(\alpha\) and \(\beta\) parameters to be between 0 and 5.

Our objective function for the optimization performed here measures the transcription error (WER in our case) of predictions made by the Whisper model, adjusted using the 5-gram model scores. For each trial, a different value of \(\alpha\) and \(\beta\)\ is suggested, and they are used to transcribe the train and validation dataset examples. It is crucial to avoid using the test split at this stage, as the model is prone to rapidly overfitting to it, leading to overly optimistic results in the optimized split. Due to the slowness of the search process, we parallelized it by loading up to 32 model instances in the same GPU at the same time and, hence, having multiple trials running simultaneously.

The optimization was conducted in over 100 trials, systematically exploring the parameter space. The number of trials is essential: making too few trials will not improve the results, and too many trials may lead to overfitting to the dataset, performing worse in new datasets, and not generalizing correctly. Also, for efficiency, all optimizations were performed using only the Tiny models, and the resulting parameters were then reused for the other model sizes. This approach of reusing the parameters for the other model sizes will probably not provide the best results in the large models. Still, in our experiments, it works well enough for an initial approach and is recommended due to the resource-intensive nature of the optimization process.

To summarize, this method will improve the fine-tuned models by adapting the Whisper model to account for linguistic characteristics captured by the language models, improving overall performance across different linguistic environments. That is to say, it ensures that the Whisper decoder's knowledge, which is general to all languages, is complemented by specific linguist statistical knowledge of the language models.

\subsection{Using Whisper with Large Language Models}

Following a similar approach to the integration of n-gram models, we extended our methodology to incorporate Large Language Models (LLMs) with Whisper, utilizing the same adjustment equation (Equation~\ref{eq:whisper_lm}). This integration aims to take advantage of the broader contextual knowledge of LLMs to further improve Whisper's performance, especially in terms of generalizability and robustness. For an efficient integration process, we made several adaptations: the range of parameter values (\(\alpha\) and \(\beta\)) was constrained to the interval [0, 3], and a subset of 4000 randomized recordings from both training and validation splits was used for a more manageable computation. This integration was performed in parallel across 7 GPUs to accelerate the optimization, which involved 100 trials, similar to the n-gram model integration. We leveraged specific LLMs optimized for particular languages: Latxa 7B v1.2 for Basque~\cite{etxaniz2024} (based on Llama 2~\cite{touvron2023_2}), Carballo Cerebras 1.3B for Galician~\cite{gamallo2024} (based on FLOR 6.3B~\cite{dalt2024}), and FLOR 6.3B for both Catalan and Spanish~\cite{dalt2024} (based on BLOOM-7.1B~\cite{workshop2023} and Chinchilla~\cite{hoffmann2022}). These models were selected due to their extensive pre-training on large, diverse corpora and their effectiveness in linguistic applications relevant to the respective languages. To integrate them, we feed partial transcripts generated by Whisper into the LLM and collect the output logits for the next token. We then apply a softmax to these logits, extract the highest log probability, and use it as the language model score. Specifically, if
\begin{equation}
  \ell = \mathrm{logits}(c)[-1],
\end{equation}
represents the final token logits after processing the partial transcript \(c\) with the LLM, then the score is given by
\begin{equation}
  P_{\text{LM}}(c) = \max \bigl(\mathrm{softmax}(\ell)\bigr),
\end{equation}
which we hypothesize serves as a reasonable proxy for the LLM’s confidence in the partial transcript. Actually, this provides a simple and effective measure of the LLM's trust in the next token. While this does not directly represent the sentence probability, we expect it to serve as a practical approximation for overall sentence reliability, helping to rescore the Whisper outputs at each decoding step.

\subsection{Corpora Leakage Analysis}

To ensure the integrity of our evaluations, particularly when assessing out-of-distribution datasets, we measured the degree of sentence overlap between our evaluation datasets and the corpora used to train the language models. This analysis helps confirm that a likely text data leakage is not skewing our results. Our n-gram language model corpora creation is detailed in Section~\ref{sec:corpora}. For the LLM language models' corpora, we have used Latxa Corpus v1.1~\cite{etxaniz2024} for Basque and CorpusNÓS~\cite{dediosflores2024} for Galician. The corpus used for the Catalan and Spanish LLM model is not publicly available, complicating the leakage assessments for these languages. For Catalan, we recreated the complete original, non-public LLM corpus by compiling a dataset that includes CATalog 1.0~\cite{giner2024}\footnote{\url{https://huggingface.co/datasets/projecte-aina/CATalog}}, CaWaC~\cite{ljubesic2014, armengolestape2021}\footnote{\url{https://zenodo.org/records/4519349}}, mC4~\cite{raffel2020}\footnote{\url{https://huggingface.co/datasets/allenai/c4}}, Wiki-40B~\cite{guo2020}\footnote{\url{https://huggingface.co/datasets/google/wiki40b}}, and a recent Wikipedia dump as of 2025-02-14\footnote{\url{https://dumps.wikimedia.org/}}. Similarly, for Spanish, we successfully recreated 98\% of the original non-public LLM corpus by compiling a dataset that includes mC4~\cite{raffel2020}, Wiki-40B~\cite{guo2020}, legal corpora~\cite{gutierrezfandino2021}\footnote{\url{https://zenodo.org/records/5495529}}, the parts available from the biomedical corpora~\cite{carrino2022}\footnote{\url{https://github.com/PlanTL-GOB-ES/lm-biomedical-clinical-es}}, and a recent Wikipedia dump as of 2025-02-14. Nevertheless, to ensure a fair comparison, all sentences from both the evaluation datasets and the language model corpora in this analysis have been normalized using the methods described in Section~\ref{sec:corpora}. Additionally, the Whisper text normalizer has been applied, removing punctuation, lowercasing, diacritic normalization, and other linguistic standardizations before sentence comparison~\cite{radford2022}.

\subsection{Ablation Study of Evaluation Parameters}

This study evaluates the impact of various evaluation options on the performance of the ASR using the Whisper vanilla models. We examine how changing specific parameters affects the WER by disabling them individually from our default configuration. To methodologically assess the impact of these parameters, we conducted a series of experiments in which each parameter varied while others remained in our selected settings. In simple terms, we disabled each parameter individually over our baseline. Furthermore, we will use the original multilingual OpenAI models, not the fine-tuned ones. This will adequately justify the default parameters used in the rest of our study.

The parameters studied here are the following:

\begin{itemize}
    \item \textbf{Beam size:} Beam search is used in decoding to consider multiple hypotheses at each step. A default beam size of 5 is utilized based on Whisper's suggestion to balance performance and computational efficiency. We will compare it by disabling the beam search, also known as greedy decoding.
    \item \textbf{Diacritics:} Taking the diacritics into account when evaluating the model can substantially alter the reported results. We ignored diacritics to homogenize language differences and possible encoding format inconsistencies between and within datasets.
    \item \textbf{Timestamps:} While timestamps are important for applications like subtitle generation, they are not the focus of our study. Instead, we assess whether enabling timestamps in the output affects the transcription accuracy.
    \item \textbf{Language:} Specifying the language of the audio can guide the model's recognition process, mainly when working with non-English languages and out-of-domain datasets. Initially, we hypothesized this parameter may gain importance for low-resource languages.
    \item \textbf{Temperature:} This parameter controls the randomness of the predictions, with a higher temperature introducing more variability into the output. By default, it has a range of values from 0.0 to 1.0, incremented by 0.2. This is known as a temperature scheduler (i.e., `(0.0, 0.2, 0.4, 0.6, 0.8, 1.0)`) and reflects a progressive increase typically used for long-sentence scenarios~\cite{radford2022}. In the context of Whisper, the temperature scheduler is used to mitigate common decoding issues with long-form audio contexts. In our particular case, we disabled it by default because we are working with sentence-level audio recordings, and we want to enable beam search for the language models. Still, we want to measure to what extent this parameter affects the scores in our datasets, just in case it is relevant to our particular scenario.
\end{itemize}

Our default configuration includes a beam size of 5, diacritics removed, timestamps excluded, the language provided to the model, and a single temperature value of 0.0.

\section{Experimental Setup}

This section presents the experimental framework used to evaluate the models, detailing the datasets, model configurations, and the experimental environment employed.

\subsection{Datasets}

This study utilizes a diverse set of datasets to evaluate the performance of the models across various languages. The datasets are selected to test both in-distribution (ID) and out-of-distribution (OOD) generalization ability.

\subsubsection{Audio Datasets}

The models are fine-tuned on the Common Voice version 13.0 dataset, which is used in the speech recognition community in recent work due to its diversity and size~\cite{gandhi2023, timmel2024}. These fine-tuned models are later evaluated on several other datasets to measure generalization across different acoustic and linguistic environments.

As for in-distribution evaluation, all models are evaluated using the Common Voice 13.0 dataset's test split to measure the ID WER.

Concerning the out-of-distribution Evaluation, for language-specific assessments, the models are tested on the following datasets:

\begin{itemize}
    \item \textbf{Basque:} Evaluated on AhoMyTTS (non-public)~\cite{sainz2012} and OpenSLR-76~\cite{kjartansson2020}. The AhoMyTTS is especially challenging because it is not in the public domain and uses unique sentences that are unavailable online. Therefore, it may serve as a control of possible leakages in the language model's corpora.
    \item \textbf{Galician:} Evaluated on OpenSLR-77~\cite{kjartansson2020}, and Google's FLEURS~\cite{conneau2022}.
    \item \textbf{Catalan:} Evaluated on OpenSLR-69~\cite{kjartansson2020}, and Google's FLEURS.
    \item \textbf{Spanish:} Evaluated on Google's FLEURS and Facebook's Multilingual LibriSpeech (MLS)~\cite{pratap2020}.
\end{itemize}

Further details of the datasets used for fine-tuning and evaluation are in Table~\ref{tab:dataset_sizes}.

\begin{table}[htp]
\centering
\begin{tabular}{llll|rr}
\hline
\textbf{Dataset} & \textbf{Short Name} & \textbf{Language} & \textbf{Split} & \multicolumn{1}{c}{\textbf{Recordings}}  & \multicolumn{1}{c}{\textbf{Hours}} \\
\hline
Common Voice 13 & CV13 & Basque   & train+validation &    17,509 &    25.9 \\
Common Voice 13 & CV13 & Galician & train+validation &    17,348 &    23.1 \\
Common Voice 13 & CV13 & Catalan  & train+validation & 1,063,345 & 1,643.1 \\
Common Voice 13 & CV13 & Spanish  & train+validation &   296,037 &   434.8 \\
\hline
Common Voice 13 &     CV13 & Basque & test & 6,591 & 10.5 \\
AhoMyTTS        & AhoMyTTS & Basque & -    &   590 &  0.8 \\
OpenSLR-76      &    SLR76 & Basque  & -    & 7,136 & 13.9 \\
Common Voice 13 &     CV13 & Galician & test & 6,546 &  9.4 \\
FLEURS          &   Fleurs & Galician & all  & 3,497 & 10.3 \\
OpenSLR-77      &   SLR77 & Galician & -    & 5,587 & 10.3 \\
Common Voice 13 &    CV13 & Catalan & test & 16,380 & 28.1 \\
FLEURS          &  Fleurs & Catalan & all  &  3,644 & 11.9 \\
OpenSLR-69      &   SLR69 & Catalan & -    &  4,240 &  9.4 \\
Common Voice 13 &    CV13 & Spanish & test & 15,708 & 26.8 \\
FLEURS          &  Fleurs & Spanish & all  &  4,112 & 13.3 \\
Multilingual LibriSpeech (MLS) & MLS & Spanish & test & 2,385 & 10.0 \\
\hline
\end{tabular}
\caption{Details of the datasets used for fine-tuning (top) and evaluation (bottom).}
\label{tab:dataset_sizes}
\end{table}

\begin{figure}[!tp]
  \centering
  \includegraphics[width=350px]{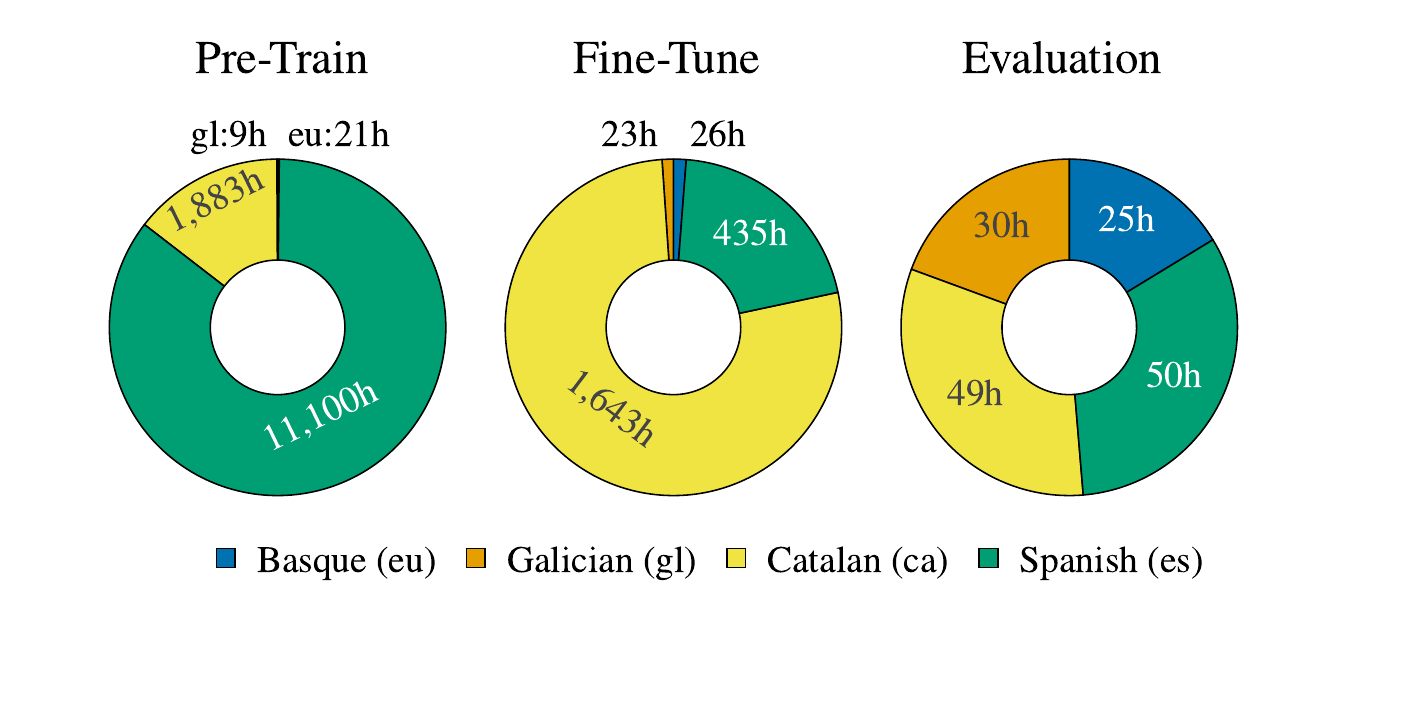}
  \caption{Distribution of dataset hours across different training phases.}
  \label{fig:data_plot}
\end{figure}

In addition to that, Figure~\ref{fig:data_plot} shows the distribution of dataset hours across different training stages: pre-training, fine-tuning, and evaluation. The pre-training hours per language, taken from the original Whisper paper~\cite{radford2022}, show Catalan's representation as medium-sized compared to other languages. During fine-tuning, Catalan is notably over-represented, surpassing even Spanish in terms of hours, which could influence model adaptability. However, for evaluation, the datasets are balanced across languages, suggesting that the precision of our final scores will likely be more equitable and unbiased across different languages. In any case, based on the data distribution shown, Basque and Galician can be considered under-resourced in this experiment.

\subsubsection{Textual Corpora for Language Models}\label{sec:corpora}

For the creation of the n-gram language models, each language's corpus is capped at 27 million sentences to maintain comparability and manageability. The main sources of these corpora are as follows: EusCrawl 1.0~\cite{artetxe2022} for Basque, SLI GalWeb Corpus~\cite{agerri2018} for Galician, Catalan Textual Corpus~\cite{ljubesic2014, armengolestape2021} for Catalan, and Spanish LibriSpeech MLS~\cite{pratap2020} for Spanish. If needed, a recent Wikipedia dump and the Opus~\cite{aulamo2019} corpus are appended to the initial corpora until it reaches 27 million sentences. Subsequently, the number of sentences for the corpus for each language is identical, making the model comparison fairer. Besides, all texts within these corpora have been segmented and normalized to ensure uniformity in language modeling processes, following established methods in the field~\cite{tyers2021}.

\subsection{Model Configurations}

This study utilizes the series of official multi-lingual Whisper models, ranging from Tiny to Large-V3, which we have fine-tuned to improve performance in specific languages. These models are available in the official OpenAI Hugging Face repository\footnote{\url{https://huggingface.co/openai}}. Their underlying structure remains consistent across models, and no internal changes were made to the original models. Basically, based on OpenAI's Whisper transformer encoder-decoder architecture~\cite{vaswani2017}, these models are adapted to handle specific languages, improving their proficiency without changing the model's constructional design.

\subsection{Experimental Environment}

The study leveraged substantial computing resources to ensure the efficient processing of models and data.

Regarding the hardware, as this experiment has been extended over time, the fine-tuning, optimization, and posterior evaluation tasks have been performed in different server instances as our infrastructure evolved. The initial fine-tuning and n-gram LM evaluations were performed on an NVIDIA A100-SXM4-80GB GPU, supported by 16 AMD EPYC\texttrademark 7513 CPU cores with 128 MB of RAM memory. For the ablation study, we used 8 AMD EPYC\texttrademark 75F3 CPU cores with a single NVIDIA A100-SXM4-80GB GPU. Lastly, all the LLM optimization and evaluation processes have been completed in 196 Intel\textregistered Xeon Platinum 8480C CPU cores with 7 NVIDIA H100 80GB HBM3 GPUs working in parallel.

As for the software, the models were developed and evaluated using a suite of software tools and libraries:

\begin{itemize}
    \item \textbf{Hugging Face library:} utilized for model fine-tuning through the scripts from the Whisper language fine-tuning community event\footref{fnot1} and for loading the LLM tokenizers and models.
    \item \textbf{Whisper ASR:} we utilized the official OpenAI's Whisper implementation for the language model integration and all evaluation tasks\footnote{\url{https://github.com/openai/whisper/} (version v20231117)}.
    \item \textbf{Common Voice Utils:} used to preprocess the corpora for the language models by segmenting and normalizing the texts in the different languages~\cite{tyers2021}\footnote{\url{https://github.com/ftyers/commonvoice-utils/}}.
    \item \textbf{KenLM library:} used to construct robust 5-gram language models trained for each language\footnote{\url{https://kheafield.com/code/kenlm/}} together with its Python package for the final integration.
    \item \textbf{Optuna framework:} for the optimization of the language model parameters\footnote{\url{https://optuna.org/}}.
    \item \textbf{JiWER:} used for measuring the CER and WER, providing accurate and reliable evaluation metrics\footnote{\url{https://jitsi.github.io/jiwer/}}.
\end{itemize}

\subsection{Evaluation Metrics}

In this study, we primarily utilize the WER~\cite{wang2003} to assess the accuracy of the fine-tuned Whisper models across different datasets and languages. The WER is a standard metric in speech recognition that measures the performance of an automatic speech recognition system. Particularly, it calculates the errors in the transcribed text, which is analogous to accuracy in more traditional classification tasks. Although WER is a widely used metric, comparing systems across diverse languages and datasets can sometimes be challenging. For this reason, in this work, we have included additional measures derived from WER to facilitate more consistent comparisons across these varied conditions.

\textbf{Relative Error Reduction (RER).}
In response to these challenges, we will employ the RER metric~\cite{zhang2010,devries2023,radford2022}, calculated using Equation~\ref{eq:rer}. This metric helps contextualize improvements by calculating the error reduction relative to a baseline, providing a simpler view of performance changes across diverse conditions. By focusing on relative rather than absolute improvements, RER also aligns better with the diminishing returns often observed in neural models as their performance increases~\cite{kaplan2020}.

\begin{equation}
\label{eq:rer}
    RER = \left(1 - \frac{WER_{\text{with intervention}}}{WER_{\text{baseline}}}\right) \times 100\%
\end{equation}

\textbf{Effective Robustness of Relative Error Reduction (ERER).} To compare the robustness of different fine-tuning and language modeling approaches in speech recognition, we adapt the concept of effective robustness from Taori et al.~\cite{taori2020} to the ASR domain, following a method similar to that in the Whisper paper~\cite{radford2022}.

In Taori et al., effective robustness is defined as in Equation~\ref{eq:taori2020}, where \(acc_1(f)\) denotes the model’s accuracy on an in-distribution (ID) test set, \(acc_2(f)\) is its accuracy on an out-of-distribution (OOD) test set under some distribution shift, and \(\xi(\cdot)\) is a baseline function that predicts \(acc_2\) given \(acc_1\). A distribution shift is understood as a change in data properties not covered by the ID test set. Since higher accuracy corresponds to better performance, a model lying exactly on the baseline has \(\rho(f) = 0\), while a model exceeding the expected OOD performance has \(\rho(f) > 0\).

\begin{equation}
\label{eq:taori2020}
    \rho(f) = acc_2(f) \;-\; \xi\bigl(acc_1(f)\bigr)
\end{equation}

With regard to ASR models, we typically use the WER metric, where lower values are better. However,  to maintain consistency with the "higher is better" paradigm from Taori et al., we instead measure RER. Specifically, let \(RER_{\mathrm{ID}}(f)\) be the relative error reduction on an ID dataset compared to a reference system, and \(RER_{\mathrm{OOD}_i}(f)\) the relative error reduction on the \(i\)-th OOD dataset. We then define ERER in Equation~\ref{eq:erer_2}, where \(N\) is the total number of OOD datasets and \(f\) denotes the model. Intuitively, \(ERER(f)\) measures how much better (or worse) a model performs under OOD conditions relative to its ID performance.

\begin{equation}
\label{eq:erer_2}
    ERER(f) \;=\; \frac{1}{N}\sum_{i=1}^{N}
    \Bigl(RER_{\mathrm{OOD}_i}(f)\;-\;RER_{\mathrm{ID}}(f)\Bigr),
\end{equation}

In practice, the function \(\xi\) from Taori et al.~\cite{taori2020} is conceptualized as the direct ID performance, assuming that the expected baseline OOD performance should ideally match the ID performance. This simplification reflects our empirical observation that OOD performance generally tracks ID performance closely but is shifted by a constant factor reflecting the generalization gap.

To clarify, the ERER intuitively captures the relative improvement or degradation in model performance when subjected to new or unseen data scenarios, providing a direct measure of a model's robustness beyond conventional accuracy metrics. An ERER value close to zero indicates that the model’s OOD performance scales with its ID performance in a predictable way, suggesting balanced robustness. Negative ERER values imply that the model underperforms on OOD data relative to what one would expect, given its ID results. Conversely, positive ERER values, though less common, imply that the model exceeds its expected performance under a distribution shift. From a robustness standpoint, the ideal scenario is an ERER of zero or above, indicating that the model’s OOD performance keeps pace with (or even surpasses) its ID performance, thereby maintaining uniformly high performance across both ID and OOD conditions.

 The RER and ERER metrics ensure that our evaluations are both exhaustive and easily understandable, facilitating a better understanding of the effectiveness of the proposed fine-tuning and language modeling interventions. Nevertheless, the detailed WER scores for each model configuration and dataset are provided in Appendix~\ref{appendix:eval_details}, allowing further inspection.

\subsection{Statistical Significance Analysis}

To evaluate the statistical significance of the differences observed in our results, we employ the Wilcoxon signed-rank test~\cite{wilcoxon1945}. This non-parametric test is ideal for our analysis as it does not assume a normal data distribution, making it well-suited for real-world datasets that may be skewed or non-uniform. The test is used to determine if the median of the differences between paired observations, such as WERs across different models or configurations, significantly deviates from zero, indicating a true effect rather than a result of random variation~\cite{Santafe2015}.

The Wilcoxon signed-rank test has been applied in two scenarios within our study:
\begin{enumerate}
    \item \textbf{Method-level significance:} To assess the overall influence of the different methodological interventions (such as fine-tuning, LM integration, or ablated parameters) on model performance. In this case, we will compare the dataset-level WER scores across methods. This analysis will help us assess the impact of the method employed.
    \item \textbf{Sentence-level significance:} To analyze the impact of individual result values across specific datasets and model sizes, ensuring a fine-grained evaluation of performance variations. In this method, we will compare individual sentence-level WERs across the different methods. These comparisons will be displayed in the results tables as a superscript in the scores.
\end{enumerate}

For statistical comparisons, we juxtapose the fine-tuned model results against the baseline vanilla models and subsequently compare the outcomes of n-gram and large language model integrations against their respective fine-tuned baselines. This approach mirrors the sequential application of each method to the models, reflecting the incremental improvements aimed at each step. Unless otherwise stated, results will be considered statistically significant at a p-value threshold of less than 0.001, often referred to as very highly significant or extremely significant. This strict criterion helps ensure that the observed differences are highly unlikely to be due to chance, thereby confirming the efficacy of the tested approaches. This methodological rigor is critical for drawing reliable conclusions from our experiments and posterior analysis.

\section{Results}

The following subsections provide and analyze the results of the fine-tuning process, the integration of language models with the fine-tuned models, the corpora leakage analysis, and, finally, the parameter ablation study of vanilla models.

\subsection{Performance Improvements}

This subsection discusses the improvements observed when integrating language models with Whisper for various languages and datasets, comparing the performance before and after the integration of language models.

\subsubsection{Fine-Tuning Results}

As shown in Table~\ref{tab:fine_tuning_impact}, fine-tuning Whisper models undoubtedly improves WER across most scenarios in low and medium-resource languages. The most prominent gains are seen in ID datasets, where improvements are as high as 76\% for the Medium and Large models in the Basque language. Notably, improvements for OOD datasets are somewhat lower but still substantial, peaking at 68\% improvement for the Medium model in Basque in the AhoMyTTS dataset. In general, the upgrades seem to affect all the model sizes, from the Tiny to Large models, with an overall positive trend. For details on the WER values from which these improvements were calculated, refer to Appendix~\ref{appendix:eval_wer}.

\begin{table}[htp]
\centering
\caption{The relative error reduction for fine-tuned models compared to vanilla models, with ID datasets listed at the top and OOD datasets at the bottom. Values indicating the highest improvements are marked in \textbf{bold}, and negative reductions are highlighted in \textcolor{red}{red}. Significance levels are indicated as follows: \(p\signa < 0.05\), \(p\signb < 0.01\), \(p\signc < 0.001\), no-superscript meaning no significance.}
\label{tab:fine_tuning_impact}
\begin{tabular}{ll|rrrrrrr}
\hline
\textbf{Language} & \textbf{Dataset} & \multicolumn{1}{c}{\textbf{Tiny}} & \multicolumn{1}{c}{\textbf{Base}} & \multicolumn{1}{c}{\textbf{Small}} & \multicolumn{1}{c}{\textbf{Medium}} & \multicolumn{1}{c}{\textbf{Large}} & \multicolumn{1}{c}{\textbf{L-V2}} & \multicolumn{1}{c}{\textbf{L-V3}} \\
\hline
Basque & CV13 & +69\%\signc & \textbf{+73\%}\signc & \textbf{+75\%}\signc & \textbf{+76\%}\signc & \textbf{+76\%}\signc & \textbf{+74\%}\signc & \textbf{+73\%}\signc \\
Galician & CV13 & +56\%\signc & +64\%\signc & +66\%\signc & +68\%\signc & +65\%\signc & +64\%\signc & +64\%\signc \\
Catalan & CV13 & \textbf{+73\%}\signc & +71\%\signc & +65\%\signc & +69\%\signc & +71\%\signc & +70\%\signc & +59\%\signc \\
Spanish & CV13 & +40\%\signc & +35\%\signc & +23\%\signc & +23\%\signc & +19\%\signc & +14\%\signc & \textcolor{red}{-2\%}\nosign \\
\hline
Basque & AhoMyTTS & \textbf{+61\%}\signc & \textbf{+66\%}\signc & \textbf{+66\%}\signc & \textbf{+68\%}\signc & \textbf{+64\%}\signc & \textbf{+60\%}\signc & \textbf{+62\%}\signc \\
Basque & SLR76 & +59\%\signc & +63\%\signc & +62\%\signc & +62\%\signc & +60\%\signc & +58\%\signc & +58\%\signc \\
Galician & Fleurs & +43\%\signc & +48\%\signc & +47\%\signc & +41\%\signc & +29\%\signc & +25\%\signc & +14\%\signc \\
Galician & SLR77 & +46\%\signc & +55\%\signc & +50\%\signc & +52\%\signc & +46\%\signc & +48\%\signc & +49\%\signc \\
Catalan & Fleurs & +48\%\signc & +39\%\signc & +11\%\signc & 0\%\signc & \textcolor{red}{-10\%}\nosign & \textcolor{red}{-24\%}\signc & \textcolor{red}{-48\%}\signc \\
Catalan & SLR69 & +57\%\signc & +50\%\signc & +32\%\signc & +32\%\signc & +28\%\signc & +21\%\signc & +8\%\signc \\
Spanish & Fleurs & \textcolor{red}{-1\%}\signc & \textcolor{red}{-6\%}\signc & \textcolor{red}{-17\%}\signc & \textcolor{red}{-9\%}\signc & \textcolor{red}{-15\%}\signc & \textcolor{red}{-10\%}\signc & \textcolor{red}{-12\%}\signc \\
Spanish & MLS & \textcolor{red}{-11\%}\signc & \textcolor{red}{-14\%}\signc & \textcolor{red}{-11\%}\signc & \textcolor{red}{-16\%}\signc & \textcolor{red}{-12\%}\signc & \textcolor{red}{-31\%}\signc & \textcolor{red}{-40\%}\signc \\
\hline
\end{tabular}
\end{table}

In contrast, the results for the high-resource language in OOD datasets are mixed. For instance, fine-tuning negatively impacts Spanish models when evaluating them with datasets like MLS, where WER worsened by up to 40\% in the Large-V3 model configuration. This trend affects smaller models less, but still, there is a slightly worsening effect, and no model size shows improvements in OOD datasets for Spanish. This probably means the pre-trained model decoder already has considerable Spanish linguistic knowledge. Consequently, during fine-tuning, it accidentally overfits to the dataset and loses the ability to generalize to other contexts.

Similarly, medium-resource languages like Catalan exhibit uneven improvements. This is particularly true in one OOD dataset, where larger models do not consistently yield better results, occasionally showing degradation in performance, as with the Large-V3 model in the FLEURS dataset. Nevertheless, the small models still benefit from fine-tuning, with a peak at 57\% in the Tiny model with the OpenSLR-69 dataset, and the net result is very positive.

Altogether, the changes produced by the fine-tuning method are statistically significant, confirming the effectiveness of fine-tuning across various languages and model sizes (\(W=261.0\), \(p=1.07 \cdot 10^{-11}\)). Moreover, most of the individual performance improvements for each dataset and model size are also statistically significant. For detailed significance levels, refer to the annotations in the Table~\ref{tab:fine_tuning_impact} (\(p\signa < 0.05\), \(p\signb < 0.01\), \(p\signc < 0.001\), no-superscript indicating no significance).

\subsubsection{N-Gram Language Model Integration Results}

In this section, an n-gram language model is integrated with the fine-tuned Whisper model. As previously stated, the LM weight optimization process was performed only using the Tiny models of each language, whose parameters were later reused for the other model sizes. Hence, the smallest model results are the most relevant for our analysis. Nonetheless, all results of all the model sizes will be reported and analyzed.

Overall, integrating language models after the fine-tuning demonstrates general positive results. Table~\ref{tab:lm_fine_tuning_impact} shows the final results when comparing the scores with the fine-tuned models without a language model. The improvement percentages listed are incremental, indicating additional gains on top of those achieved through fine-tuning alone. For instance, a +37\% improvement in the Tiny model for Basque means a further 37\% error reduction beyond the initial 69\% improvement from fine-tuning. The improvement is compared with fine-tuned models as a baseline, even for the cases where the models do not clearly improve, like in Spanish. In other words, we added the language models to the fine-tuned models in all cases, even if they were not clearly better than the vanilla models. Notwithstanding these results, if the goal is to obtain the best performance, attaching the language model to the original models might be a better approach for high-resource languages (see Appendix~\ref{appendix:eval_vanilla_es} for further details).

\begin{table}[htp]
\centering
\caption{The relative error reduction for fine-tuned models with a 5-gram language model compared with the previous fine-tuned results without the language model. ID datasets are listed at the top, and OOD datasets are listed at the bottom. Values indicating the highest improvements are marked in \textbf{bold}, and negative reductions are highlighted in \textcolor{red}{red}. Significance levels are indicated as follows: \(p\signa < 0.05\), \(p\signb < 0.01\), \(p\signc < 0.001\), no-superscript meaning no significance.}
\label{tab:lm_fine_tuning_impact}
\begin{tabular}{ll|rrrrrrr}
\hline
\textbf{Language} & \textbf{Dataset} & \multicolumn{1}{c}{\textbf{Tiny}} & \multicolumn{1}{c}{\textbf{Base}} & \multicolumn{1}{c}{\textbf{Small}} & \multicolumn{1}{c}{\textbf{Medium}} & \multicolumn{1}{c}{\textbf{Large}} & \multicolumn{1}{c}{\textbf{L-V2}} & \multicolumn{1}{c}{\textbf{L-V3}} \\
\hline
Basque & CV13 & \textbf{+37\%}\signc & \textbf{+45\%}\signc & \textbf{+50\%}\signc & \textbf{+50\%}\signc & \textbf{+49\%}\signc & \textbf{+48\%}\signc & \textbf{+51\%}\signc \\
Galician & CV13 & +32\%\signc & +39\%\signc & +40\%\signc & +39\%\signc & +35\%\signc & +33\%\signc & +38\%\signc \\
Catalan & CV13 & +14\%\signc & +14\%\signc & +11\%\signc & 0\%\signc & +4\%\signc & +7\%\signc & +7\%\signc \\
Spanish & CV13 & +22\%\signc & +24\%\signc & +25\%\signc & +23\%\signc & +19\%\signc & +19\%\signc & +18\%\signc \\
\hline
Basque & AhoMyTTS & +9\%\signc & \textbf{+29\%}\signc & \textbf{+23\%}\signc & \textbf{+34\%}\signc & \textbf{+30\%}\signc & \textbf{+31\%}\signc & \textbf{+30\%}\signc \\
Basque & SLR76 & \textbf{+21\%}\signc & +22\%\signc & +22\%\signc & +22\%\signc & +20\%\signc & +20\%\signc & +23\%\signc \\
Galician & Fleurs & +14\%\signc & +15\%\signc & +8\%\signc & \textcolor{red}{-3\%}\signa & \textcolor{red}{-2\%}\signc & \textcolor{red}{-1\%}\signc & \textcolor{red}{-4\%}\nosign \\
Galician & SLR77 & +21\%\signc & +20\%\signc & +20\%\signc & +16\%\signc & +16\%\signc & +15\%\signc & +14\%\signc \\
Catalan & Fleurs & +13\%\signc & +12\%\signc & +12\%\signc & +5\%\signc & \textcolor{red}{-1\%}\signc & +1\%\signc & +7\%\signc \\
Catalan & SLR69 & +19\%\signc & +18\%\signc & +19\%\signc & +13\%\signc & +13\%\signc & +13\%\signc & +12\%\signc \\
Spanish & Fleurs & +7\%\signc & +5\%\signc & \textcolor{red}{-8\%}\nosign & +1\%\signc & 0\%\signc & 0\%\signb & \textcolor{red}{-1\%}\signb \\
Spanish & MLS & +4\%\signc & +9\%\signc & \textcolor{red}{-2\%}\signc & \textcolor{red}{-9\%}\signc & \textcolor{red}{-5\%}\signc & \textcolor{red}{-17\%}\nosign & \textcolor{red}{-10\%}\nosign \\
\hline
\end{tabular}
\end{table}

The results show remarkable improvements in ID datasets, with the Large-V3 model in the Basque Common Voice dataset improving by up to 51\%. The improvements for OOD datasets reach up to 34\%, as observed in the Medium model for the Basque language in the FLEURS dataset.

Contrary to fine-tuning results, adding language models still benefits the smaller models of high-resource languages, such as Spanish, or, alternatively, it helps mitigate some of the performance declines observed during fine-tuning. Nonetheless, as it happens with fine-tuning, it still does not consistently benefit large models. For example, when tested on Spanish OOD datasets like FLEURS and MLS, the Large, Large-V2, and Large-V3 models often show minimal to negative improvements. This could be indicative of the limitations of the current LM parameters used, which were optimized based on the Tiny models and may not scale correctly across larger model architectures. Additionally, the quality of corpora used for training the language models might also be influencing these scores.

With regard to the statistical effect of n-grams integration, the differences are very highly significant (\(W=186.0\), \(p=9.95 \cdot 10^{-13}\)), demonstrating the value of this method. Most of the individual score changes are also statistically significant, as shown in Tables~\ref{tab:lm_fine_tuning_impact}'s superscripts, with only a few exceptions in high-resource or large models.

To summarize, smaller models benefit from LM integration in all the cases tested here, suggesting that language models can raise performance where fine-tuning alone may not suffice.

\subsubsection{Large Language Model Integration Results}

With respect to the integration of large language models with the fine-tuned Whisper models, the scores consistently show performance improvements across all tested languages and datasets, albeit with more modest gains compared to n-gram LM integrations. Table~\ref{tab:llm_wer_improvement} displays the relative error reductions achieved by adding LLMs to the fine-tuned models.

\begin{table}[htp]
\centering
\caption{The relative error reduction for fine-tuned models with a large language model compared with the fine-tuned results without any language model. ID datasets are listed at the top, and OOD datasets are listed at the bottom. Values indicating the biggest improvements are marked in \textbf{bold}, and values that improved over 5-gram LM results are \underline{underlined}. Significance levels are indicated as follows: \(p\signa < 0.05\), \(p\signb < 0.01\), \(p\signc < 0.001\), no-superscript meaning no significance.}
\label{tab:llm_wer_improvement}
\begin{tabular}{ll|rrrrrrr}
\hline
\textbf{Language} & \textbf{Dataset} & \multicolumn{1}{c}{\textbf{Tiny}} & \multicolumn{1}{c}{\textbf{Base}} & \multicolumn{1}{c}{\textbf{Small}} & \multicolumn{1}{c}{\textbf{Medium}} & \multicolumn{1}{c}{\textbf{Large}} & \multicolumn{1}{c}{\textbf{L-V2}} & \multicolumn{1}{c}{\textbf{L-V3}} \\
\hline
Basque & CV13 & \textbf{+18\%}\signc & \textbf{+20\%}\signc & \textbf{+23\%}\signc & \textbf{+19\%}\signc & \textbf{+16\%}\signc & \textbf{+16\%}\signc & \textbf{+16\%}\signc \\
Galician & CV13 & +8\%\signc & +9\%\signc & +7\%\signc & +6\%\signc & +1\%\signc & +3\%\signc & +4\%\signc \\
Catalan & CV13 & +6\%\signc & +6\%\signc & +5\%\signc & \underline{+1\%}\signc & +3\%\signc & +4\%\signc & +3\%\signc \\
Spanish & CV13 & +7\%\signc & +8\%\signc & +6\%\signc & +6\%\signc & +5\%\signc & +4\%\signc & +4\%\signc \\
\hline
Basque & AhoMyTTS & \textbf{\underline{+18\%}}\signc & \textbf{+17\%}\signc & \textbf{+19\%}\signc & \textbf{+17\%}\signc & \textbf{+16\%}\signb & \textbf{+16\%}\signb & \textbf{+15\%}\signa \\
Basque & SLR76 & +15\%\signc & +15\%\signc & +15\%\signc & +12\%\signc & +11\%\signc & +11\%\signc & +11\%\signc \\
Galician & Fleurs & +9\%\signc & +8\%\signc & +7\%\signc & \underline{+5\%}\signc & \underline{+4\%}\signc & \underline{+3\%}\signc & \underline{+2\%}\signc \\
Galician & SLR77  & +9\%\signc & +8\%\signc & +7\%\signc & +4\%\signc & +5\%\signc & +4\%\signc & +7\%\signc \\
Catalan & Fleurs & +9\%\signc & +9\%\signc & +7\%\signc & +5\%\signc & \underline{+4\%}\signc & \underline{+5\%}\signc & +4\%\signc \\
Catalan & SLR69  & +8\%\signc & +10\%\signc & +8\%\signc & +5\%\signc & +4\%\signc & +7\%\signc & +4\%\signc \\
Spanish & Fleurs & +4\%\signc & +4\%\signc & \underline{+2\%}\signb & \underline{+1\%}\signc & \underline{+2\%}\signc & \underline{+1\%}\signa & \underline{0\%}\signb \\
Spanish & MLS    & \underline{+8\%}\signc & \underline{+12\%}\signc & \underline{+9\%}\signc & \underline{+3\%}\signc & \underline{+8\%}\signc & \underline{+7\%}\signc & \underline{+2\%}\signc \\
\hline
\end{tabular}
\end{table}

The improvements were more prominent in the Basque language, where the Small model size reduced WER by up to 23\% in the ID dataset. This superior performance in Basque could be attributed to the Latxa model's robust adaptation to the language, possibly improved by its extensive pre-training on a large and adequately curated Basque corpus. Although the overall improvements are lower than those seen with n-gram LMs, this initial integration of LLMs demonstrates promising directions for improving ASR systems for low-resource languages.

Expanding on this, the addition of LLMs improved performance across both ID and OOD datasets much more evenly than adding n-gram LMs did. In some cases, the improvements in OOD datasets slightly surpassed those in ID, indicating strong generalization ability. For instance, the Basque language models showed consistent improvements, with OOD datasets such as AhoMyTTS and OpenSLR-76 exhibiting improvements pretty close to those in the ID dataset.

As with the previous methods, the integration of large language models also results in statistically significant differences in performance (\(W=0.0\), \(p=1.71 \cdot 10^{-15}\)). In this case, all the performance improvements achieved through LLM integration are statistically significant at least at the \(p\signa < 0.05\) level, with most scores exhibiting very high statistical significance, indicated by \(p\signc < 0.001\). This consistency in statistical significance emphasizes the robustness of LLM integration across different model configurations and datasets.

Moreover, in contrast to the first methods, where high-resource languages such as Spanish showed some divergent or negative trends with fine-tuning and n-gram LM addition, the integration of LLMs into these models resulted in consistent improvements. This suggests that LLMs, with their broader and deeper understanding of language, can help overcome some of the overfitting issues seen before in fine-tuning and n-gram LM integrations.

\subsubsection{Comparison of Method Robustness}

In this subsection, we further analyze the robustness of the various methods employed.

\begin{figure}[!tp]
  \centering
  \begin{minipage}{0.48\textwidth}
    \centering
    \includegraphics[width=\textwidth]{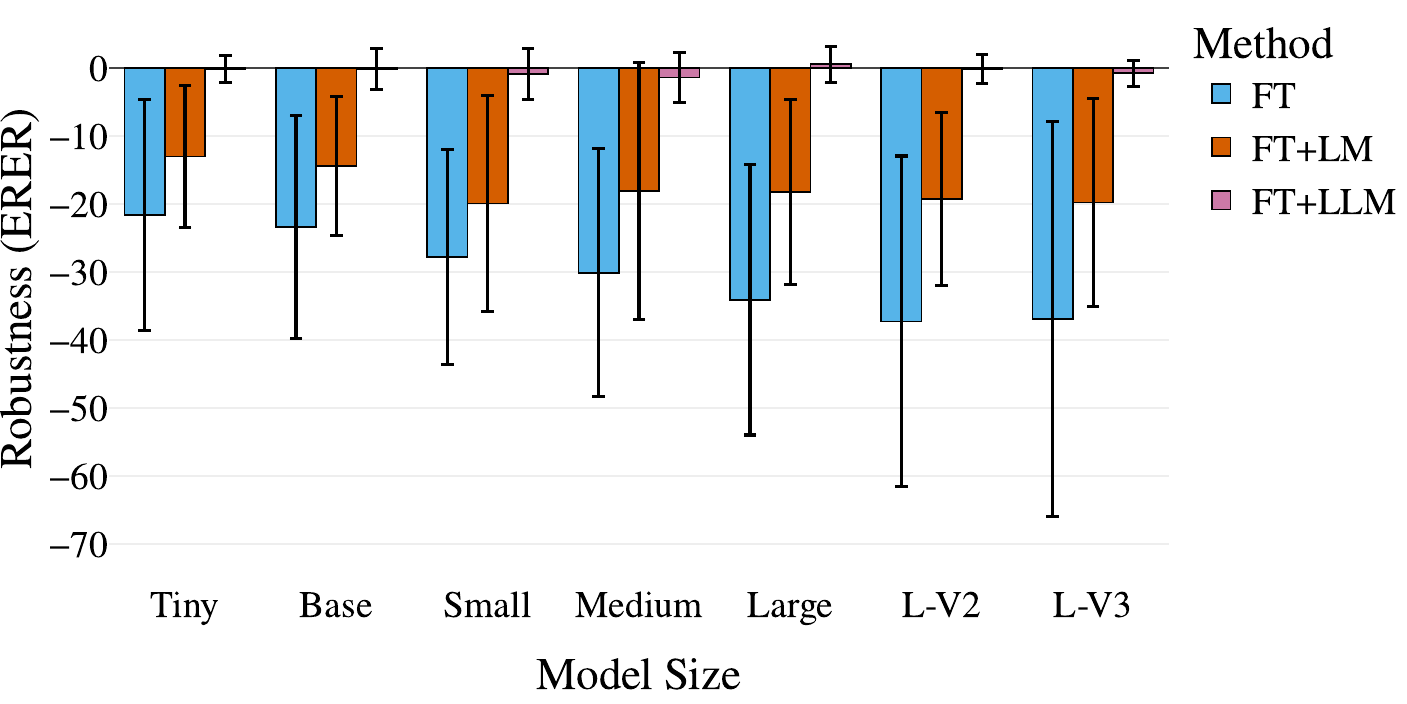}
    \caption{Effective robustness of RER by model size.}
    \label{fig:erer_by_size_method}
  \end{minipage}\hfill
  \begin{minipage}{0.48\textwidth}
    \centering
    \includegraphics[width=\textwidth]{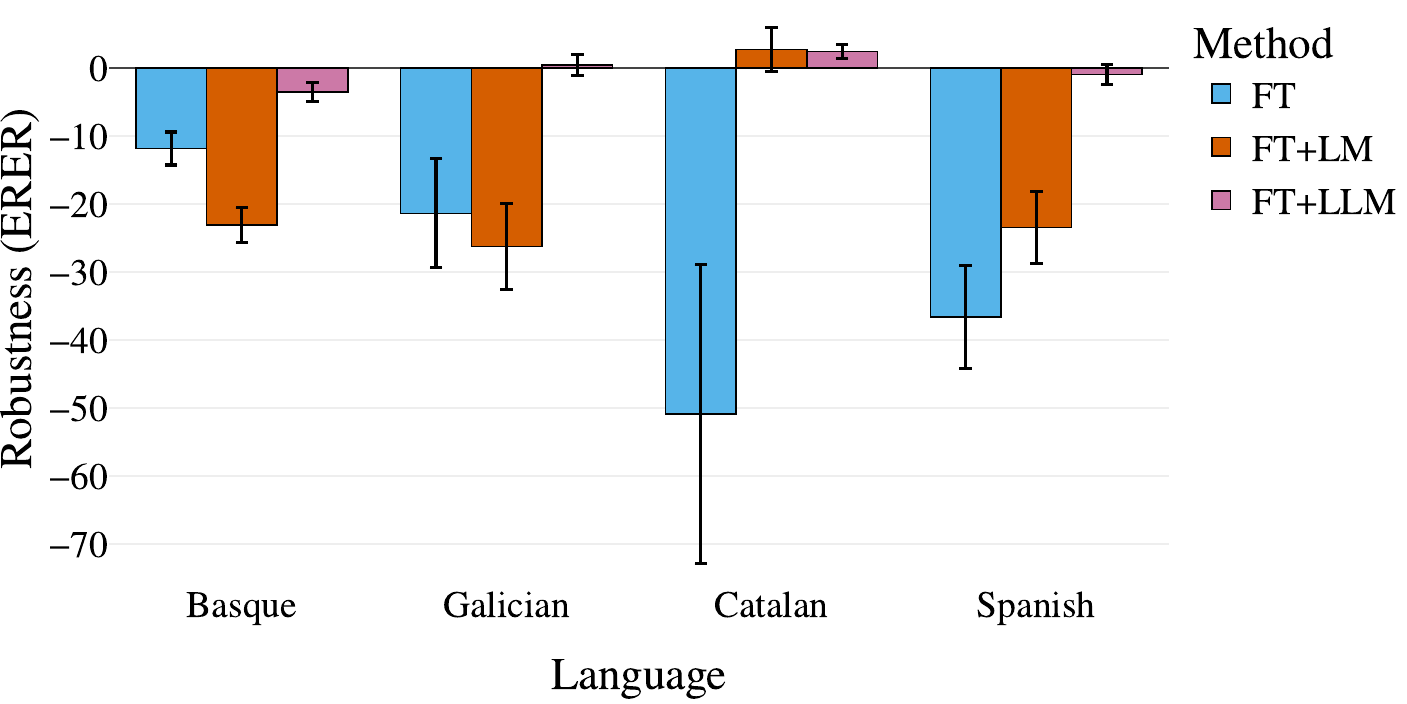}
    \caption{Effective robustness of RER by language.}
    \label{fig:erer_by_method_lang}
  \end{minipage}
\end{figure}

Figure~\ref{fig:erer_by_size_method} provides a visual analysis of the effective robustness of RER (ERER) by model size. It reveals a trend where increasing model size tends to decrease robustness when only fine-tuning is applied, indicating a risk of dataset overfitting. This tendency toward decreasing robustness, even though much more subtle, is also present in the LM results. However, this trend does not persist with the addition of LLMs, which uniformly improve robustness across all model sizes.

Figure~\ref{fig:erer_by_method_lang} extends this analysis by comparing methods across different languages. It illustrates that while the transition from fine-tuning alone to integrating LMs does not uniformly improve robustness across languages, as in Basque and Galician, the addition of LLMs consistently elevates robustness, notably outperforming other methods. This suggests the greater effectiveness of LLMs, which appear to better mitigate language-specific challenges. Interestingly, Catalan, which demonstrated lower robustness during initial fine-tuning, shows significant improvement upon the addition of LMs.

Unlike the variations observed with fine-tuning and n-gram LMs, LLMs contribute to more stable improvements across different testing scenarios. The effective robustness analysis confirms that LLM integration not only does increase the baseline performance but does so consistently across varied environments and contexts. These insights highlight the considerable variability between languages and emphasize the importance of considering model size when developing robust ASR systems.

As we continue to refine the integration techniques and as the LLMs themselves evolve, we anticipate further improvements in ASR performance, particularly for languages that have traditionally been underrepresented in speech technology resources. The promising results in both ID and OOD settings highlight the capacity of LLMs to contribute to the robustness and accuracy of multilingual models.

\subsection{Results of Corpora Leakage Analysis}

Regarding the results from our data leakage analysis presented in Table~\ref{tab:dataset_leaks}, sentence overlap in the ID datasets is notably high, exceeding 75\% for most languages in the n-gram language models and ranging from 18\% to 84\% for the large language models. This high overlap in Common Voice datasets is partly due to the inclusion of sentences from Wikipedia in the recording prompts\footnote{\url{https://common-voice.github.io/community-playbook/sub_pages/text.html}}, which is also a common source for corpora used in language model training. The Common Voice Catalan dataset presents a notable exception, showing lower overlap due to their proactive efforts to generate genuine, diverse sentences through collaborations with authors, publishers, and public entities, avoiding reliance on web-crawled data~\cite{armentanooller2024}. In contrast to the ID datasets, the OOD datasets exhibit minimal to no leakage (less than 15\%), suggesting that the robustness of the LLMs is not merely a result of memorization. Curiously, using TTS datasets for ASR evaluations proves valuable, as they overlap less with today's language model corpora. Overall, these findings reinforce our confidence that the observed improvements in OOD datasets are genuine and not influenced by memorized sentences.

\begin{table}[htp]
\centering
\caption{Percentage of sentences from evaluation datasets found in the training corpora of n-gram and large language models.}
\label{tab:dataset_leaks}
\begin{tabular}{ll|rr}
\hline
\textbf{Language} & \textbf{Dataset} & \multicolumn{1}{c}{\textbf{LM Corpus}} & \multicolumn{1}{c}{\textbf{LLM Corpus}} \\
\hline
Basque    & CV13     & 79.82\% & 36.34\% \\
Galician  & CV13     & 79.04\% & 36.40\% \\
Catalan   & CV13     &  7.39\% & 17.53\% \\
Spanish   & CV13     & 77.09\% & 83.87\% \\
\hline
Basque    & AhoMyTTS & 13.39\% & 11.19\% \\
Basque    & SLR76    &  0.25\% &  0.27\% \\
Galician  & Fleurs   &  0.00\% &  0.00\% \\
Galician  & SLR77    &  0.88\% &  0.81\% \\
Catalan   & Fleurs   &  0.00\% &  0.00\% \\
Catalan   & SLR69    &  0.33\% &  0.92\% \\
Spanish   & Fleurs   &  0.00\% &  0.00\% \\
Spanish   & MLS      &  0.00\% &  4.23\% \\
\hline
\end{tabular}
\end{table}

\subsection{Ablation Study of the Impact of Evaluation Parameters}

The ablation study focuses on understanding the impact of various evaluation parameters on the WER across different languages and Whisper model sizes.

\begin{figure}[!tp]
  \centering
  \includegraphics[width=450px]{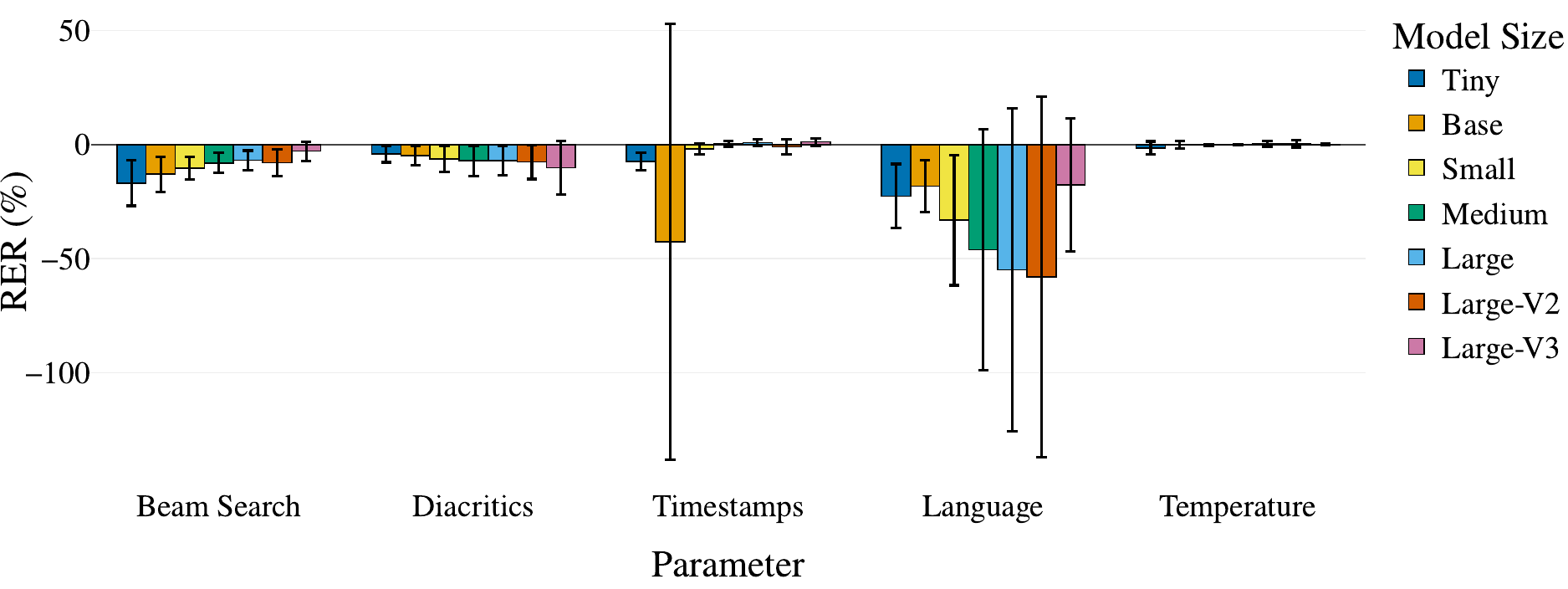}
  \caption{The averaged RER across different model sizes to study the impact of various evaluation parameters on the WER. Negative values indicate performance decreases when changing from our selected baseline.}
  \label{fig:ablation_results}
\end{figure}

Figure~\ref{fig:ablation_results} provides an overview of how the removal or change of specific parameters affects the WER across all languages and datasets. Each bar in the plot displays the mean RER value and its standard deviation, with negative values indicating a deterioration in performance and positive values indicating an improvement over the established baseline configuration: a beam size of 5, diacritics removed, timestamps excluded, the language provided to the model, and a temperature of 0. Detailed WER values used for these calculations can be found in Appendix~\ref{appendix:eval_ablation}.

The statistical test results revealed significant differences for most parameters, as follows:

The removal of the language specification has the biggest impact, worsening the WER by as much as 58\% for larger model configurations, with significant statistical evidence (\(W=0.0\), \(p=5.36 \cdot 10^{-15}\)). This substantial degradation highlights the importance of specifying the language to maintain high transcription accuracy and indicates the model's difficulties in recognizing languages with lower resources. Likewise, disabling beam search results in an average decrease in WER ranging from 3\% to 17\%, with strong statistical support for these findings (\(W=6.0\), \(p=2.12 \cdot 10^{-15}\)), underscoring the effectiveness of beam search in enhancing model precision across various sizes.

As expected, incorporating diacritics during evaluation consistently has a negative impact on WER, a change that is statistically significant (\(W=0.0\), \(p=1.71 \cdot 10^{-15}\)). That said, disabling them might not always be a sensible decision, depending on specific use cases. Meanwhile, including timestamps shows a variable impact with an overall trend of slightly decreasing performance, depending on the model size and specific dataset conditions. This effect, too, is statistically significant (\(W=857.0\), \(p=3.49 \cdot 10^{-5}\)). This variability indicates that including timestamp information may negatively impact the scores, especially for the smaller models, and it is advisable to disable it if not needed. For example, the Base model shows a particular sensitivity to this parameter. During the initial pre-training of the models, the timestamped examples were probably more scarce and may have biased the model slightly to produce worse results.

Interestingly, disabling the temperature scheduler results in negligible changes to the WER across any model size, with most of the RER values remaining at or near 0\%. Indeed, this parameter does not statistically influence the WER (\(W=680.5\), \(p=0.59\)), indicating an insignificant impact on performance across different model sizes and conditions. This is particularly relevant to our study as it supports our decision to disable this feature when integrating the language models. The absence of changes due to the temperature scheduler can be attributed to the nature of our datasets, which consist of short sentences. It is important to note that the impact of this parameter might vary in scenarios involving longer or more complex audio recordings, where temperature control could help mitigate repetitive or overly simplistic generation patterns often seen in longer sequences.

As a result, our chosen parameter configuration demonstrates the best overall performance, as deviations from this setup typically result in increased WERs.

\section{Discussion}

This study has demonstrated that the integration of statistical language models with the Whisper system can remarkably reduce word error rates across a variety of languages and datasets. Particularly in minority languages, these improvements suggest that while Whisper's decoder is phonetically robust, it may not sufficiently capture the full range of linguistic characteristics necessary for optimal performance without additional help. This is evident from the substantial improvements observed when language models are integrated, highlighting a shortcoming in the ability of the model's native decoder to function effectively as a language model, particularly at the grammatical levels of low-resource languages.

Furthermore, the variability in improvements, especially in high-resource languages and larger model architectures, points to a complex interaction between model size, language resources, and the parameters of the language model. The observed dependency of the language models' effectiveness on the model size and dataset characteristics underscores the importance of the context in applying these improvements. Remarkably, the weighting \(\alpha\) and \(\beta\) parameters for the language model integration appear to have some relationship with the language characteristics or the dataset resources used, as the good results obtained by their reuse imply. In particular, our evaluations indicate that the Large-V3 models consistently outperform other versions, with the Large-V2 also showing superior performance than Large-V1. Accordingly, we recommend the use of Large-V3 for those seeking the most accurate Whisper model, addressing the concerns raised in recent discussions\footnote{\url{https://deepgram.com/learn/whisper-v3-results}}.

Alongside these findings, the integration of LLMs with Whisper, while giving more modest improvements, demonstrates a clear increase in robustness. This robustness, along with expected future improvements as LLMs are better adapted to specific languages, indicates that they could play an important role in evolving ASR systems towards greater linguistic generalizability and reliability. Currently, there seems to exist a trade-off between using traditional language models for performance and employing LLMs for stronger robustness. In environments where specific use cases dominate or where performance optimization is critical, traditional LMs may still be preferable. Conversely, for broader, more generalized applications, especially with high-resource languages or larger models where statistical LMs struggle to improve the results, LLMs start to give promising results. Moreover, the validation provided by the corpora leakage analysis further strengthens our confidence in these improvements. It ensures that the improvements we observe are due to effective model integration, not simply a result of overfitting or memorization.

On the other hand, the ablation study conducted as part of this research has also underlined the critical nature of parameter settings in model evaluation. Adjusting these parameters can lead to considerable variances in model performance, indicating Whisper's sensitivity to configuration settings. This highlights an essential aspect of deploying Whisper in diverse linguistic settings: the configuration used to evaluate model performance needs to be consistent and well-considered to ensure reliability in the results.

\section{Conclusion and Future Work}

The findings from this study contribute to the aged but growing body of knowledge on the application of traditional language models in the field of speech recognition, particularly in enhancing the Whisper ASR models across diverse linguistic environments. We have shown that language model integration can improve performance, particularly in low-resource languages where Whisper's native decoding power may be insufficient. Although, despite the overall positive trend, the effectiveness of these models varies for the largest models with high-resource languages, indicating a need for more customized approaches to language model integration.

The introduction of large language models into the Whisper framework has also highlighted a trade-off between immediate performance boosts and the robustness of model outputs. While n-gram language models offer better improvements, LLMs provide a more stable and consistent improvement across different datasets, including in-distribution and out-of-distribution scenarios. This suggests that LLMs may be particularly valuable in environments where generalization and stability are priorities. We also hypothesize that LLMs may shine in processing long-form content, leveraging their broader contextual understanding to capture more complex language patterns than n-gram models can provide. An obvious next step would be to investigate the combined use of n-gram and large language models to conceivably merge the capabilities of both methods.

Looking ahead, it is essential to explore the optimized use of language models further, mainly how these adjustments affect different languages and model sizes. Experimenting with a broader range of parameter and hyperparameter values adjusted to specific linguistic and acoustic characteristics may produce further improvements in model performance. Moreover, expanding the scope of these improvements to include other types of language models and decoding strategies, like end-to-end approaches, could provide deeper insights into the optimal configurations for various use cases. Finally, while this work focused on a limited set of languages and datasets due to time and resource constraints, evaluating these methods across a wider range of languages, out-of-domain conditions, and acoustic models remains an important next step to confirm broader applicability.

This research proposes an initial approach to using language models in modern speech recognition technologies. It involves bringing back traditional, widely tested integration techniques and applying them to new, advanced models. By continuing to refine these approaches and adapting the integration strategies as models evolve, it may be possible to improve further the accuracy and reliability of novel systems like Whisper across all languages, not just those well-represented in training data.

\section{Acknowledgments}

The authors express their gratitude to both the DIPC Supercomputing Center and the technological management body of the Basque Government, EJIE, for their technical and human support. Additionally, this research was partially financed by the Spanish Ministry for Digital Transformation and of Civil Service and the EU-funded NextGenerationEU Recovery, Transformation, and Resilience Plan (ILENIA project, 2022/TL22/00215335). We also express our gratitude to Carlos Domínguez for his constant assistance, personal feedback, and ideas for further research, and to Andrés Piñeiro-Martín, whose insights unknowingly planted the initial seed for this project idea.

\printbibliography

\newpage

\appendix

\section{Detailed Performance Results of Whisper Models}\label{appendix:eval_details}

\subsection{Vanilla, Fine-Tuned, and Language Model WER Scores}\label{appendix:eval_wer}

\begin{table}[H]
\centering
\caption{WER scores for Whisper vanilla models without any modifications or fine-tuning.}
\label{tab:vanilla_wer_scores}
\begin{tabular}{ll|rrrrrrr}
\hline
\textbf{Language} & \textbf{Dataset} & \multicolumn{1}{c}{\textbf{Tiny}} & \multicolumn{1}{c}{\textbf{Base}} & \multicolumn{1}{c}{\textbf{Small}} & \multicolumn{1}{c}{\textbf{Medium}} & \multicolumn{1}{c}{\textbf{Large}} & \multicolumn{1}{c}{\textbf{L-V2}} & \multicolumn{1}{c}{\textbf{L-V3}} \\
\hline
  Basque &     CV13 & 97.93 & 92.86 & 71.72 &  58.27 & 50.84 &    43.20 &    38.85 \\
Galician &     CV13 & 51.48 & 46.31 & 29.37 &  20.31 & 17.84 &    15.26 &    12.46 \\
 Catalan &     CV13 & 52.94 & 41.48 & 25.22 &  17.48 & 15.98 &    14.94 &    13.67 \\
 Spanish &     CV13 & 27.69 & 18.39 &  9.70 &   6.38 &  5.81 &     5.16 &     4.38 \\
 \hline
  Basque & AhoMyTTS & 91.65 & 84.91 & 67.66 &  56.81 & 48.20 &    41.16 &    36.67 \\
  Basque &    SLR76 & 93.55 & 91.82 & 72.59 &  61.72 & 55.01 &    48.87 &    45.92 \\
Galician &   Fleurs & 48.04 & 40.53 & 24.94 &  16.48 & 14.87 &    12.41 &    10.06 \\
Galician &    SLR77 & 52.12 & 46.75 & 31.76 &  24.26 & 21.28 &    19.77 &    17.20 \\
 Catalan &   Fleurs & 40.74 & 27.68 & 14.20 &   8.57 &  7.42 &     6.18 &     5.68 \\
 Catalan &    SLR69 & 41.26 & 29.95 & 16.21 &  10.38 &  9.09 &     7.90 &     7.86 \\
 Spanish &   Fleurs & 27.70 & 22.02 & 17.46 &  15.78 & 15.44 &    15.15 &    15.01 \\
 Spanish &      MLS & 17.33 & 11.45 &  6.58 &   4.62 &  4.34 &     3.71 &     2.92 \\
\hline
\end{tabular}
\end{table}

\begin{table}[H]
\centering
\caption{WER scores of fine-tuned models.}
\label{tab:fine_tuning_wer_scores}
\begin{tabular}{ll|rrrrrrr}
\hline
\textbf{Language} & \textbf{Dataset}  & \multicolumn{1}{c}{\textbf{Tiny}} & \multicolumn{1}{c}{\textbf{Base}} & \multicolumn{1}{c}{\textbf{Small}} & \multicolumn{1}{c}{\textbf{Medium}} & \multicolumn{1}{c}{\textbf{Large}} & \multicolumn{1}{c}{\textbf{L-V2}} & \multicolumn{1}{c}{\textbf{L-V3}} \\
\hline
  Basque &     CV13 & 30.26 & 24.82 & 17.72 &  14.03 & 12.11 &    11.26 &    10.52 \\
Galician &     CV13 & 22.69 & 16.71 &  9.94 &   6.44 &  6.27 &     5.50 &     4.55 \\
 Catalan &     CV13 & 14.55 & 11.85 &  8.78 &   5.40 &  4.69 &     4.52 &     5.56 \\
 Spanish &     CV13 & 16.61 & 11.90 &  7.32 &   4.92 &  4.69 &     4.43 &     4.46 \\
\hline
  Basque & AhoMyTTS & 35.57 & 29.17 & 23.12 &  18.22 & 17.59 &    16.62 &    13.96 \\
  Basque &    SLR76 & 38.55 & 34.36 & 27.51 &  23.19 & 22.15 &    20.54 &    19.14 \\
Galician &   Fleurs & 27.41 & 21.00 & 13.24 &   9.68 & 10.60 &     9.26 &     8.61 \\
Galician &    SLR77 & 27.96 & 21.06 & 15.75 &  11.55 & 11.50 &    10.37 &     8.81 \\
 Catalan &   Fleurs & 21.32 & 16.81 & 12.68 &   8.58 &  8.16 &     7.69 &     8.38 \\
 Catalan &    SLR69 & 17.63 & 15.04 & 11.04 &   7.01 &  6.55 &     6.26 &     7.20 \\
 Spanish &   Fleurs & 28.02 & 23.26 & 20.41 &  17.25 & 17.69 &    16.61 &    16.87 \\
 Spanish &      MLS & 19.25 & 13.04 &  7.28 &   5.34 &  4.86 &     4.87 &     4.08 \\
\hline
\end{tabular}
\end{table}

\begin{table}[H]
\centering
\caption{WER scores of fine-tuned models with the n-gram language model.}
\label{tab:fine_tuning_lm_wer_scores}
\begin{tabular}{ll|rrrrrrr}
\hline
\textbf{Language} & \textbf{Dataset}  & \multicolumn{1}{c}{\textbf{Tiny}} & \multicolumn{1}{c}{\textbf{Base}} & \multicolumn{1}{c}{\textbf{Small}} & \multicolumn{1}{c}{\textbf{Medium}} & \multicolumn{1}{c}{\textbf{Large}} & \multicolumn{1}{c}{\textbf{L-V2}} & \multicolumn{1}{c}{\textbf{L-V3}} \\
\hline
  Basque &     CV13 & 18.99 & 13.70 &  8.84 &   7.07 &  6.17 &     5.88 &     5.15 \\
Galician &     CV13 & 15.34 & 10.16 &  5.97 &   3.90 &  4.05 &     3.68 &     2.80 \\
 Catalan &     CV13 & 12.54 & 10.18 &  7.77 &   5.41 &  4.49 &     4.19 &     5.16 \\
 Spanish &     CV13 & 12.94 &  8.99 &  5.50 &   3.77 &  3.78 &     3.57 &     3.67 \\
\hline
  Basque & AhoMyTTS & 32.38 & 20.69 & 17.77 &  12.09 & 12.35 &    11.49 &     9.74 \\
  Basque &    SLR76 & 30.59 & 26.76 & 21.38 &  18.01 & 17.66 &    16.43 &    14.83 \\
Galician &   Fleurs & 23.59 & 17.94 & 12.21 &   9.95 & 10.82 &     9.38 &     8.94 \\
Galician &    SLR77 & 22.20 & 16.79 & 12.62 &   9.69 &  9.70 &     8.80 &     7.59 \\
 Catalan &   Fleurs & 18.56 & 14.82 & 11.21 &   8.13 &  8.23 &     7.65 &     7.83 \\
 Catalan &    SLR69 & 14.35 & 12.38 &  8.96 &   6.09 &  5.71 &     5.45 &     6.32 \\
 Spanish &   Fleurs & 25.93 & 22.14 & 22.01 &  17.15 & 17.76 &    16.67 &    17.01 \\
 Spanish &      MLS & 18.50 & 11.81 &  7.44 &   5.81 &  5.09 &     5.69 &     4.49 \\
\hline
\end{tabular}
\end{table}

\begin{table}[H]
\centering
\caption{WER scores of fine-tuned models with the large language model.}
\label{tab:fine_tuning_llm_wer_scores}
\begin{tabular}{ll|rrrrrrr}
\hline
\textbf{Language} & \textbf{Dataset} & \multicolumn{1}{c}{\textbf{Tiny}} & \multicolumn{1}{c}{\textbf{Base}} & \multicolumn{1}{c}{\textbf{Small}} & \multicolumn{1}{c}{\textbf{Medium}} & \multicolumn{1}{c}{\textbf{Large}} & \multicolumn{1}{c}{\textbf{L-V2}} & \multicolumn{1}{c}{\textbf{L-V3}} \\
\hline
  Basque &     CV13 & 24.73 & 19.96 & 13.60 &  11.40 & 10.15 &     9.43 &     8.87 \\
Galician &     CV13 & 20.77 & 15.20 &  9.24 &   6.08 &  6.21 &     5.31 &     4.37 \\
 Catalan &     CV13 & 13.70 & 11.14 &  8.37 &   5.34 &  4.55 &     4.35 &     5.38 \\
 Spanish &     CV13 & 15.41 & 10.98 &  6.89 &   4.65 &  4.48 &     4.26 &     4.30 \\
\hline
  Basque & AhoMyTTS & 29.28 & 24.14 & 18.77 &  15.18 & 14.85 &    14.02 &    11.84 \\
  Basque &    SLR76 & 32.77 & 29.25 & 23.27 &  20.51 & 19.68 &    18.29 &    17.01 \\
Galician &   Fleurs & 25.01 & 19.25 & 12.27 &   9.21 & 10.16 &     8.97 &     8.40 \\
Galician &    SLR77 & 25.58 & 19.41 & 14.62 &  11.07 & 10.97 &     9.91 &     8.20 \\
 Catalan &   Fleurs & 19.39 & 15.30 & 11.85 &   8.15 &  7.82 &     7.29 &     8.02 \\
 Catalan &    SLR69 & 16.24 & 13.51 & 10.14 &   6.67 &  6.26 &     5.85 &     6.92 \\
 Spanish &   Fleurs & 26.89 & 22.32 & 19.99 &  17.03 & 17.34 &    16.45 &    16.81 \\
 Spanish &      MLS & 17.71 & 11.51 &  6.60 &   5.20 &  4.49 &     4.54 &     3.98 \\
\hline
\end{tabular}
\end{table}

\subsection{Language Model Integration for Vanilla Spanish Models}\label{appendix:eval_vanilla_es}

Scores in Table~\ref{tab:lm_vanilla_impact} reflect the performance of vanilla Whisper models integrated with an n-gram language model directly, without fine-tuning. The results generally are similar to those reported in Table~\ref{tab:lm_fine_tuning_impact}, with some improvements and declines being more pronounced. For further details, Table~\ref{tab:lm_wer_scores_es} shows the WER scores of the vanilla models with the language model.

\begin{table}[H]
\centering
\caption{The relative error reduction for vanilla Spanish models with a 5-gram language model compared with the original vanilla results without the language model. Negative reductions are highlighted in \textcolor{red}{red}.}
\label{tab:lm_vanilla_impact}
\begin{tabular}{ll|rrrrrrr}
\hline
\textbf{Language} & \textbf{Dataset} & \multicolumn{1}{c}{\textbf{Tiny}} & \multicolumn{1}{c}{\textbf{Base}} & \multicolumn{1}{c}{\textbf{Small}} & \multicolumn{1}{c}{\textbf{Medium}} & \multicolumn{1}{c}{\textbf{Large}} & \multicolumn{1}{c}{\textbf{L-V2}} & \multicolumn{1}{c}{\textbf{L-V3}} \\
\hline
Spanish & CV13 & +16\% & +20\% & +23\% & +22\% & +22\% & +20\% & +25\% \\
Spanish & Fleurs & +6\% & +5\% & +1\% & \textcolor{red}{-3\%} & \textcolor{red}{-4\%} & \textcolor{red}{-1\%}  & \textcolor{red}{-7\%} \\
Spanish & MLS & +9\% & +3\% & \textcolor{red}{-16\%} & \textcolor{red}{-63\%} & \textcolor{red}{-52\%} & \textcolor{red}{-29\%} & \textcolor{red}{-47\%} \\
\hline
\end{tabular}
\end{table}

\begin{table}[H]
\centering
\caption{WER scores of vanilla with the n-gram language model for Spanish.}
\label{tab:lm_wer_scores_es}
\begin{tabular}{ll|rrrrrrr}
\hline
\textbf{Language} & \textbf{Dataset}  & \multicolumn{1}{c}{\textbf{Tiny}} & \multicolumn{1}{c}{\textbf{Base}} & \multicolumn{1}{c}{\textbf{Small}} & \multicolumn{1}{c}{\textbf{Medium}} & \multicolumn{1}{c}{\textbf{Large}} & \multicolumn{1}{c}{\textbf{L-V2}} & \multicolumn{1}{c}{\textbf{L-V3}} \\
\hline
\textit{Spanish} & \textit{CV13} & 23.35 & 14.63 & 7.52 & 4.96 & 4.53 & 4.14 & 3.30 \\
Spanish & Fleurs & 26.08 & 20.92 & 17.35 & 16.29 & 16.04 & 15.24 & 16.04 \\
Spanish & MLS & 15.85 & 11.16 & 7.61 & 7.54 & 6.58 & 4.80 & 4.28 \\
\hline
\end{tabular}
\end{table}

\subsection{Ablation Study WER Scores}\label{appendix:eval_ablation}

\begin{table}[H]
\centering
\caption{WER scores in the ablation study results for the Basque language.}
\label{tab:ablation_detailed_eu}
\begin{tabular}{ll|rrrrrrr}
\hline
\textbf{Parameter} & \textbf{Dataset} & \multicolumn{1}{c}{\textbf{Tiny}} & \multicolumn{1}{c}{\textbf{Base}} & \multicolumn{1}{c}{\textbf{Small}} & \multicolumn{1}{c}{\textbf{Medium}} & \multicolumn{1}{c}{\textbf{Large}} & \multicolumn{1}{c}{\textbf{L-V2}} & \multicolumn{1}{c}{\textbf{L-V3}} \\
\hline
\textit{Baseline} & \textit{CV13} & \textit{97.93} & \textit{92.86} & \textit{71.72} & \textit{58.27} & \textit{50.84} & \textit{43.20} & \textit{38.85} \\
No Beam Size & CV13 & 105.52 & 98.45 & 83.90 & 65.70 & 57.92 & 50.29 & 42.98 \\
Diacritics & CV13 & 98.03 & 92.93 & 71.83 & 58.32 & 50.93 & 43.26 & 38.90 \\
Timestamps & CV13 & 101.67 & 98.75 & 71.72 & 58.09 & 50.47 & 43.44 & 39.10 \\
No Language & CV13 & 136.73 & 116.79 & 112.40 & 93.08 & 72.92 & 68.22 & 41.67 \\
Temp. Scheduler & CV13 & 105.31 & 96.03 & 72.17 & 58.30 & 50.82 & 43.21 & 38.87 \\
\hline
\textit{Baseline} & \textit{AhoMyTTS} & \textit{91.65} & \textit{84.91} & \textit{67.66} & \textit{56.81} & \textit{48.20} & \textit{41.16} & \textit{36.67} \\
No Beam Size & AhoMyTTS & 103.68 & 92.90 & 73.78 & 64.11 & 54.78 & 45.27 & 39.59 \\
Diacritics & AhoMyTTS & 91.97 & 85.64 & 68.14 & 57.67 & 48.90 & 41.94 & 37.27 \\
Timestamps & AhoMyTTS & 100.66 & 87.81 & 67.56 & 56.40 & 47.81 & 41.33 & 36.54 \\
No Language & AhoMyTTS & 118.92 & 99.55 & 87.24 & 73.77 & 56.52 & 46.43 & 37.13 \\
Temp. Scheduler & AhoMyTTS & 93.63 & 84.94 & 67.66 & 56.81 & 48.20 & 41.16 & 36.67 \\
\hline
\textit{Baseline} & \textit{SLR76} & \textit{93.55} & \textit{91.82} & \textit{72.59} & \textit{61.72} & \textit{55.01} & \textit{48.87} & \textit{45.92} \\
No Beam Size & SLR76 & 101.32 & 97.20 & 78.16 & 67.34 & 59.85 & 54.34 & 48.20 \\
Diacritics & SLR76 & 93.77 & 92.12 & 72.97 & 62.08 & 55.40 & 49.21 & 46.28 \\
Timestamps & SLR76 & 98.85 & 98.46 & 72.65 & 62.00 & 55.00 & 48.85 & 46.15 \\
No Language & SLR76 & 126.77 & 103.16 & 96.94 & 80.50 & 64.62 & 57.74 & 47.58 \\
Temp. Scheduler & SLR76 & 95.15 & 90.22 & 72.43 & 61.67 & 55.01 & 48.87 & 45.92 \\
\hline
\end{tabular}
\end{table}

\begin{table}[H]
\centering
\caption{WER scores in the ablation study results for the Galician language.}
\label{tab:ablation_detailed_gl}
\begin{tabular}{ll|rrrrrrr}
\hline
\textbf{Parameter} & \multicolumn{1}{c}{\textbf{Dataset}} & \multicolumn{1}{c}{\textbf{Tiny}} & \multicolumn{1}{c}{\textbf{Base}} & \multicolumn{1}{c}{\textbf{Small}} & \multicolumn{1}{c}{\textbf{Medium}} & \multicolumn{1}{c}{\textbf{Large}} & \multicolumn{1}{c}{\textbf{L-V2}} & \multicolumn{1}{c}{\textbf{L-V3}} \\
\hline
\textit{Baseline} & \textit{CV13} & \textit{51.48} & \textit{46.31} & \textit{29.37} & \textit{20.31} & \textit{17.84} & \textit{15.26} & \textit{12.46} \\
No Beam Size & CV13 & 59.72 & 53.33 & 31.91 & 21.61 & 19.19 & 16.60 & 12.86 \\
Diacritics & CV13 & 55.66 & 50.51 & 32.90 & 23.03 & 20.10 & 17.53 & 14.64 \\
Timestamps & CV13 & 54.85 & 49.13 & 30.08 & 20.33 & 17.99 & 15.25 & 12.36 \\
No Language & CV13 & 66.79 & 60.08 & 51.36 & 50.33 & 50.96 & 48.26 & 22.47 \\
Temp. Scheduler & CV13 & 51.21 & 45.74 & 29.46 & 20.39 & 17.87 & 15.27 & 12.46 \\
\hline
\textit{Baseline} & \textit{Fleurs} & \textit{48.04} & \textit{40.53} & \textit{24.94} & \textit{16.48} & \textit{14.87} & \textit{12.41} & \textit{10.06} \\
No Beam Size & Fleurs & 54.33 & 43.68 & 26.97 & 17.54 & 15.73 & 13.25 & 10.26 \\
Diacritics & Fleurs & 51.77 & 44.26 & 28.14 & 18.96 & 17.03 & 14.47 & 12.10 \\
Timestamps & Fleurs & 49.03 & 41.45 & 25.08 & 16.68 & 14.72 & 12.38 & 10.06 \\
No Language & Fleurs & 58.05 & 51.41 & 43.72 & 38.18 & 40.49 & 33.35 & 13.72 \\
Temp. Scheduler & Fleurs & 46.36 & 40.17 & 24.95 & 16.48 & 14.87 & 12.41 & 10.06 \\
\hline
\textit{Baseline} & \textit{SLR77} & \textit{52.12} & \textit{46.75} & \textit{31.76} & \textit{24.26} & \textit{21.28} & \textit{19.77} & \textit{17.20} \\
No Beam Size & SLR77 & 59.15 & 51.74 & 34.43 & 25.48 & 22.44 & 21.16 & 17.30 \\
Diacritics & SLR77 & 55.53 & 50.22 & 34.48 & 26.41 & 22.93 & 21.40 & 18.81 \\
Timestamps & SLR77 & 54.46 & 49.21 & 31.98 & 24.06 & 21.06 & 19.60 & 16.79 \\
No Language & SLR77 & 64.57 & 59.62 & 47.66 & 41.57 & 46.69 & 39.78 & 21.73 \\
Temp. Scheduler & SLR77 & 51.88 & 46.67 & 31.77 & 24.27 & 21.28 & 19.77 & 17.20 \\
\hline
\end{tabular}
\end{table}

\begin{table}[H]
\centering
\caption{WER scores in the ablation study results for the Catalan language.}
\label{tab:ablation_detailed_ca}
\begin{tabular}{ll|rrrrrrr}
\hline
\textbf{Parameter} & \multicolumn{1}{c}{\textbf{Dataset}} & \multicolumn{1}{c}{\textbf{Tiny}} & \multicolumn{1}{c}{\textbf{Base}} & \multicolumn{1}{c}{\textbf{Small}} & \multicolumn{1}{c}{\textbf{Medium}} & \multicolumn{1}{c}{\textbf{Large}} & \multicolumn{1}{c}{\textbf{L-V2}} & \multicolumn{1}{c}{\textbf{L-V3}} \\
\hline
\textit{Baseline} & \textit{CV13} & \textit{52.94} & \textit{41.48} & \textit{25.22} & \textit{17.48} & \textit{15.98} & \textit{14.94} & \textit{13.67} \\
No Beam Size & CV13 & 65.85 & 50.82 & 29.59 & 19.29 & 17.33 & 16.43 & 14.33 \\
Diacritics & CV13 & 54.13 & 42.53 & 25.94 & 18.06 & 16.46 & 15.35 & 14.11 \\
Timestamps & CV13 & 57.82 & 45.92 & 26.03 & 17.80 & 15.86 & 15.28 & 13.75 \\
No Language & CV13 & 74.81 & 58.13 & 35.63 & 27.18 & 23.88 & 24.51 & 16.59 \\
Temp. Scheduler & CV13 & 52.03 & 41.10 & 25.23 & 17.45 & 15.98 & 14.94 & 13.67 \\
\hline
\textit{Baseline} & \textit{Fleurs} & \textit{40.74} & \textit{27.68} & \textit{14.20} & \textit{8.57} & \textit{7.42} & \textit{6.18} & \textit{5.68} \\
No Beam Size & Fleurs & 51.04 & 34.07 & 16.52 & 9.22 & 8.01 & 6.77 & 5.95 \\
Diacritics & Fleurs & 42.08 & 28.80 & 14.81 & 8.97 & 7.77 & 6.45 & 5.96 \\
Timestamps & Fleurs & 42.90 & 29.30 & 14.51 & 8.39 & 7.37 & 6.27 & 5.67 \\
No Language & Fleurs & 57.32 & 40.61 & 26.86 & 15.05 & 15.36 & 10.24 & 6.97 \\
Temp. Scheduler & Fleurs & 40.16 & 27.17 & 14.20 & 8.57 & 7.40 & 6.18 & 5.68 \\
\hline
\textit{Baseline} & \textit{SLR69} & \textit{41.26} & \textit{29.95} & \textit{16.21} & \textit{10.38} & \textit{9.09} & \textit{7.90} & \textit{7.86} \\
No Beam Size & SLR69 & 51.58 & 35.80 & 18.55 & 11.11 & 9.53 & 8.63 & 8.02 \\
Diacritics & SLR69 & 42.66 & 31.08 & 17.06 & 10.96 & 9.67 & 8.34 & 8.39 \\
Timestamps & SLR69 & 42.76 & 31.13 & 16.28 & 10.34 & 8.89 & 8.03 & 7.79 \\
No Language & SLR69 & 54.55 & 39.15 & 18.86 & 13.71 & 11.38 & 10.41 & 8.93 \\
Temp. Scheduler & SLR69 & 41.25 & 29.53 & 16.21 & 10.38 & 9.09 & 7.90 & 7.83 \\
\hline
\end{tabular}
\end{table}

\begin{table}[H]
\centering
\caption{WER scores in the ablation study results for the Spanish language.}
\label{tab:ablation_detailed_es}
\begin{tabular}{ll|rrrrrrr}
\hline
\textbf{Parameter} & \multicolumn{1}{c}{\textbf{Dataset}} & \multicolumn{1}{c}{\textbf{Tiny}} & \multicolumn{1}{c}{\textbf{Base}} & \multicolumn{1}{c}{\textbf{Small}} & \multicolumn{1}{c}{\textbf{Medium}} & \multicolumn{1}{c}{\textbf{Large}} & \multicolumn{1}{c}{\textbf{L-V2}} & \multicolumn{1}{c}{\textbf{L-V3}} \\
\hline
\textit{Baseline} & \textit{CV13} & \textit{27.69} & \textit{18.39} & \textit{9.70} & \textit{6.38} & \textit{5.81} & \textit{5.16} & \textit{4.38} \\
No Beam Size & CV13 & 38.01 & 23.54 & 11.31 & 6.99 & 6.31 & 5.72 & 4.52 \\
Diacritics & CV13 & 28.67 & 19.23 & 10.33 & 6.93 & 6.36 & 5.63 & 4.82 \\
Timestamps & CV13 & 31.27 & 20.23 & 9.86 & 6.47 & 5.85 & 5.42 & 4.30 \\
No Language & CV13 & 29.94 & 19.74 & 10.31 & 6.88 & 6.24 & 5.53 & 4.60 \\
Temp. Scheduler & CV13 & 28.25 & 18.56 & 9.76 & 6.40 & 5.83 & 5.21 & 4.38 \\
\hline
\textit{Baseline} & \textit{Fleurs} & \textit{27.70} & \textit{22.02} & \textit{17.46} & \textit{15.78} & \textit{15.44} & \textit{15.15} & \textit{15.01} \\
No Beam Size & Fleurs & 32.04 & 24.05 & 18.03 & 15.87 & 15.56 & 15.32 & 14.95 \\
Diacritics & Fleurs & 28.30 & 22.44 & 17.74 & 15.97 & 15.63 & 15.33 & 15.14 \\
Timestamps & Fleurs & 31.25 & 79.01 & 17.64 & 15.71 & 15.06 & 14.59 & 14.94 \\
No Language & Fleurs & 33.19 & 27.37 & 18.77 & 15.78 & 15.44 & 15.15 & 15.02 \\
Temp. Scheduler & Fleurs & 27.70 & 22.02 & 17.46 & 15.78 & 15.44 & 15.15 & 14.96 \\
\hline
\textit{Baseline} & \textit{MLS} & \textit{17.33} & \textit{11.45} & \textit{6.58} & \textit{4.62} & \textit{4.34} & \textit{3.71} & \textit{2.92} \\
No Beam Size & MLS & 20.73 & 13.26 & 7.24 & 5.22 & 4.48 & 3.70 & 2.85 \\
Diacritics & MLS & 18.81 & 12.67 & 7.53 & 5.41 & 5.08 & 4.42 & 3.86 \\
Timestamps & MLS & 18.20 & 11.96 & 7.05 & 4.48 & 4.22 & 3.93 & 2.80 \\
No Language & MLS & 17.53 & 11.53 & 6.77 & 4.88 & 4.79 & 3.90 & 2.93 \\
Temp. Scheduler & MLS & 17.33 & 11.45 & 6.58 & 4.62 & 4.21 & 3.56 & 2.89 \\
\hline
\end{tabular}
\end{table}

\newpage

\subsubsection{Optimization of Parameter Value Ranges in Language Model Integration}

\begin{figure}[!tp]
  \centering
  \begin{minipage}{0.41\textwidth}
    \centering
    \includegraphics[width=\textwidth]{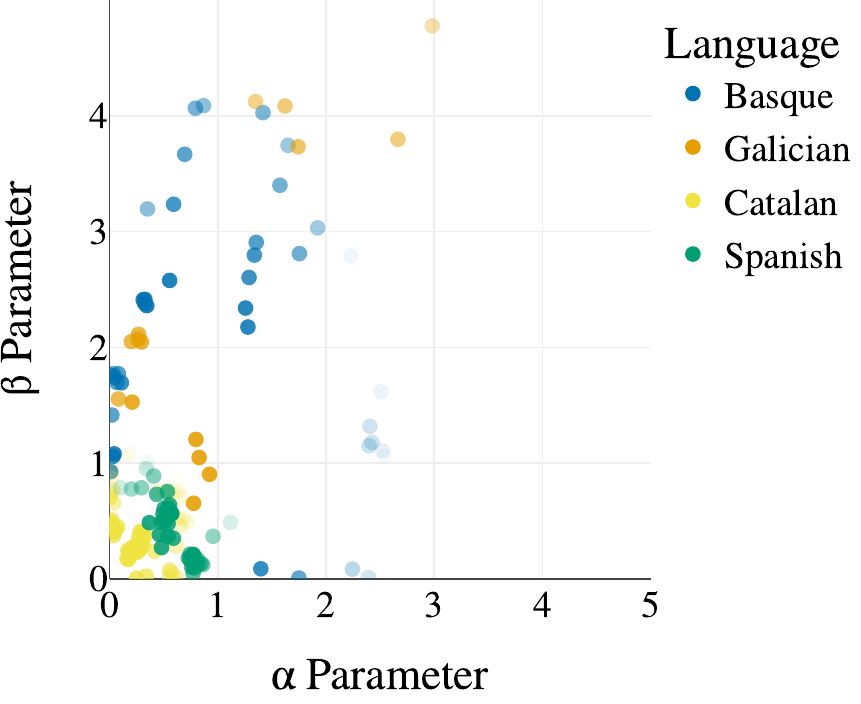}
    \caption{LM optimization trials with better scores being more opaque.}
    \label{fig:opt_lm}
  \end{minipage}\hfill
  \begin{minipage}{0.41\textwidth}
    \centering
    \includegraphics[width=\textwidth]{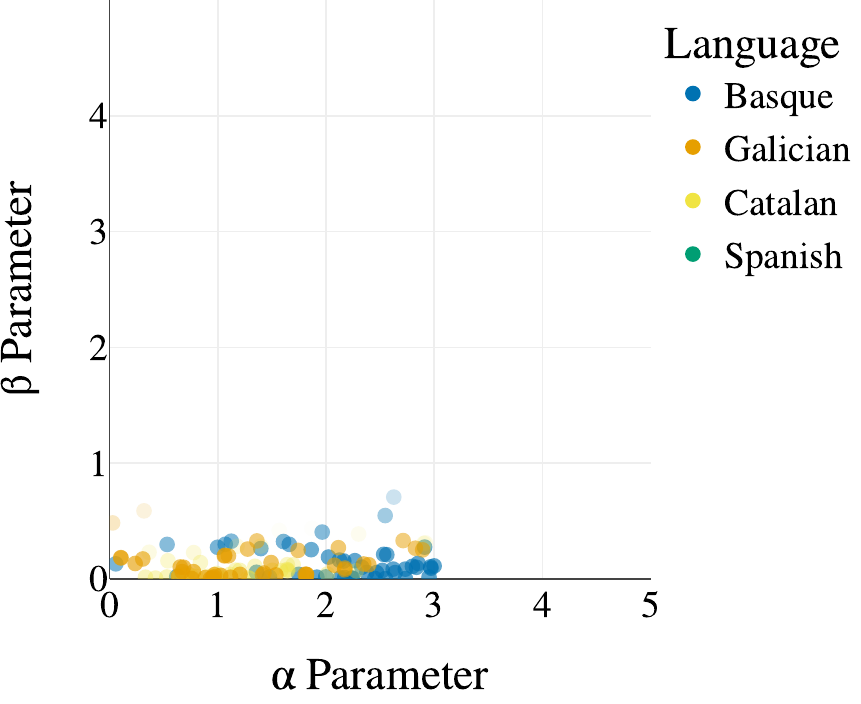}
    \caption{LLM optimization trials with better scores being more opaque.}
    \label{fig:opt_llm}
  \end{minipage}
\end{figure}

In this section, we present the optimization trials for the weighting parameters used to integrate the n-gram and large language models with the acoustic model results. Figures~\ref{fig:opt_lm} and~\ref{fig:opt_llm} illustrate these optimization processes by mapping the \(\alpha\) and \(\beta\) values against the resulting WER for each trial in the validation split. The opacity of each marker represents its corresponding WER score: markers with a very low WER are more opaque, indicating better performance, while markers with a WER approaching 15\% are nearly transparent, denoting lower performance.

In Figure~\ref{fig:opt_lm}, we observe that trials for lower-resource languages often present higher \(\beta\) values, indicating a greater reliance on sentence length adjustments within the n-gram model framework. These languages are more widely spread across the plot, suggesting a diverse range of effective parameter configurations. In contrast, higher-resource languages are clustered around lower parameter values, typically within the \([0..1]\) range for both \(\alpha\) and \(\beta\), indicating less variability in their effective optimization settings.

Figure~\ref{fig:opt_llm} reveals an outstanding trend where the \(\beta\) parameter generally holds much lower values or appears to be nearly irrelevant. This pattern suggests that, due to their extensive context understanding, large language models do not benefit as much from modifications based on sentence length as their n-gram counterparts do. This implies that their intrinsic knowledge may already account for contextual length internally, rendering additional length-based adjustments unnecessary.

The results of these optimization trials show that some characteristics of the languages may influence the parameter values. We hope the observed value ranges will serve as a useful benchmark for refining the optimization process for additional languages in the future. However, since our study was limited to only four languages, our ability to generalize these findings across a broader linguistic spectrum remains constrained.

\end{document}